\documentclass[11pt, a4paper, copyright]{google}

\usepackage[numbers,compress]{natbib}
\usepackage{hyperref}       
\definecolor{linkred}{HTML}{a33a32}
\hypersetup{
  colorlinks=true,
  linkcolor=linkred,
  citecolor=linkred,
  urlcolor=linkred,
  pdfborder={0 0 0}
}
\usepackage{nicefrac}       
\usepackage[most]{tcolorbox}

\newcommand{\answerTODO}[1][]{\textcolor{red}{\bf [TODO]}}

\uselogo{}
\correspondingauthor{}
\renewcommand{\copyrightext}{{\footerfont\itshape \textsuperscript{*}Equal contribution. Corresponding author(s): yh0068@princeton.edu}}

\makeatletter
\renewcommand{\absfont}{\normalfont\linespread{1.2}\fontsize{11}{12}\selectfont}
\renewcommand{\titlefont}{\color{black}\normalfont\bfseries\fontsize{18.5}{23}\selectfont}
\makeatother

\title{Toward Skill-Native LLMs: Skill Entropy for Benchmarking and Training Long-Horizon Reasoning}

\author[1]{Yinghui He\textsuperscript{*}}
\author[1]{Ling Yang\textsuperscript{*}}
\author[2]{Jiarui Liu}
\author[3]{Yongjin Yang}
\author[4]{Lechen Zhang}
\author[5]{Yingcheng Wu}
\author[6]{Zhenfei Yin}
\author[1]{Mengdi Wang}
\author[1]{Sanjeev Arora}

\affil[1]{Princeton University}
\affil[2]{Carnegie Mellon University}
\affil[3]{University of Toronto}
\affil[4]{University of Illinois Urbana-Champaign}
\affil[5]{Stanford University}
\affil[6]{University of Oxford}

\usepackage{amsmath,amsfonts,bm}

\usepackage{cleveref}
\usepackage{wrapfig}
\usepackage{caption}

\usepackage[most]{tcolorbox}
\usepackage[table]{xcolor}
\usepackage{listings}
\usepackage[T1]{fontenc}
\usepackage{subcaption}
\usepackage{makecell}
\usepackage{multirow}
\usepackage{twemojis}

\def\eqref#1{equation~\ref{#1}}

\def\1{\bm{1}}

\DeclareMathAlphabet{\mathsfit}{\encodingdefault}{\sfdefault}{m}{sl}
\SetMathAlphabet{\mathsfit}{bold}{\encodingdefault}{\sfdefault}{bx}{n}

\definecolor{lightpink}{HTML}{f7ded8}
\definecolor{lightyellow}{HTML}{f7eede}
\definecolor{lightgreen}{HTML}{DDEBDD}

\definecolor{pink}{HTML}{d85a4a}
\definecolor{yellow}{HTML}{c6a15b}
\definecolor{green}{HTML}{789c80}
\definecolor{red}{HTML}{9A3F3A}

\newcommand{\skerl}{\text{Skill-Entropy RL}}
\newcommand{\bench}{\text{Skill\textsuperscript{2}-Bench}}

\newcommand{\single}{{\textcolor{yellow}{\(\scalebox{1.25}{$\boldsymbol{\bullet}$}\)}}}
\newcommand{\cross}{\textbf{\textcolor{yellow}{\(\scalebox{1.25}{$\boldsymbol{\circ}$}\!\rule[0.5ex]{0.5em}{0.35pt}\!\scalebox{1.25}{$\boldsymbol{\bullet}$}\!\rule[0.5ex]{0.5em}{0.35pt}\!\scalebox{1.25}{$\boldsymbol{\circ}$}\)}}}
\newcommand{\cc}{\cellcolor{gray!10}}

\newcommand{\qwen}{Qwen3-4B-Instruct}
\newcommand{\Sqwen}{Qwen3-1.7B}

\newtcolorbox{prompt}[3][]{
    colback=#2,
    colbacktitle=#3,
    colframe=#3,
    coltitle=white,
    fontupper=\small\ttfamily,
    fonttitle=\small\bfseries,
    boxsep=2pt,
    left=0pt,
    right=0pt,
    top=0pt,
    bottom=0pt,
    boxrule=0.8pt,
    enhanced,
    breakable,
    #1,
}

\newtcblisting{promptlisting}[2][]{
    enhanced,
    breakable,
    colback=lightgreen!50,
    colbacktitle=green,
    colframe=green,
    coltitle=white,
    fonttitle=\small\bfseries,
    boxsep=2pt,
    left=4pt,
    right=4pt,
    top=2pt,
    bottom=2pt,
    boxrule=0.6pt,
    listing only,
    listing options={
        basicstyle=\scriptsize\ttfamily,
        breaklines=true,
        breakatwhitespace=false,
        columns=fullflexible,
        keepspaces=true,
        showstringspaces=false,
        breakindent=0pt,
    },
    title={#2},
    #1,
}

\newtcblisting{examplelisting}[2][]{
    enhanced,
    breakable,
    colback=lightyellow!50,
    colbacktitle=yellow!85,
    colframe=yellow!85,
    coltitle=white,
    fonttitle=\small\bfseries,
    boxsep=2pt,
    left=4pt,
    right=4pt,
    top=2pt,
    bottom=2pt,
    boxrule=0.6pt,
    listing only,
    listing options={
        basicstyle=\scriptsize\ttfamily,
        breaklines=true,
        breakatwhitespace=false,
        columns=fullflexible,
        keepspaces=true,
        showstringspaces=false,
        breakindent=0pt,
    },
    title={#2},
    #1,
}

\begin{abstract}
\vspace{-10pt}
{\fontsize{11pt}{11pt} \selectfont \raisebox{-0.06em}{\includegraphics[height=1em]{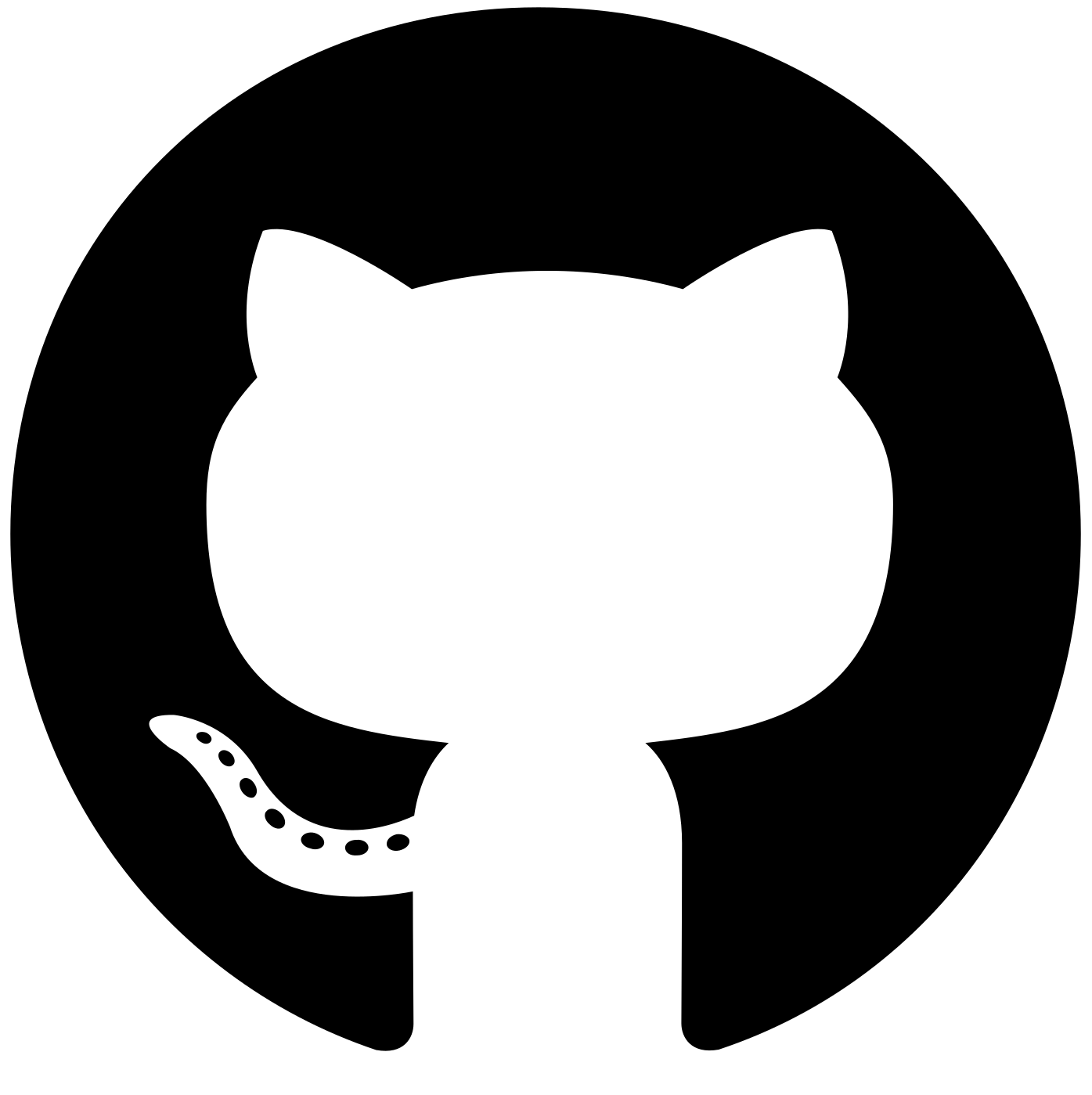}} \href{https://github.com/Gen-Verse/Skill-Entropy-RL}{https://github.com/Gen-Verse/Skill-Entropy-RL} \quad{\includegraphics[height=1em]{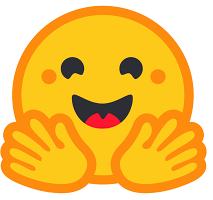}} \href{https://huggingface.co/datasets/Gen-Verse/Skill2-Bench}{Skill$^2$-Bench}}
\\
\\
\looseness-1 Long-horizon reasoning in recent LLMs demands that the model switch between distinct skills inside a reasoning chain, such as first doing a math derivation, then using the result to plan a schedule. We call such problems \emph{cross-skill long-horizon tasks}: multi-step tasks whose steps require different reasoning skills and depend on earlier outputs. Existing benchmarks often evaluate individual skills, lacking a principled way to measure \emph{how well a model switches between skills}.

We address this gap from both the evaluation and training sides. We introduce \textbf{\emph{Skill Entropy}}, a measure of the difficulty of switching from one skill to another. We then propose \textbf{\bench{}}, a benchmark of cross-skill long-horizon tasks built over $558$ skills across 9 verifiable and open-ended domains. Each task is assigned a task-level skill-entropy score and grouped into three difficulty levels. Evaluating 8 frontier and 4 open-source models on \bench{} reveals skill-switching gap: accuracy decreases on higher-entropy tasks.
We then turn skill entropy from a benchmark scale into a training signal.
We propose \textbf{Skill-Entropy RL}, an RL framework where the model predicts not only the answer at each step but also the skill used to produce it. The reward combines step-level correctness with a \emph{skill-entropy reward} that measures the alignment between the model-predicted skill sequence and the gold skill sequence.
On \qwen{} and \Sqwen{}, Skill-Entropy RL improves the \bench{} score from $34.4\%$ to $68.4\%$ and from $14.6\%$ to $40.1\%$ respectively, outperforming competitive baselines. The same pipeline can be applied to off-the-shelf training data such as OpenR1-Math, indicating that skill entropy is a reusable training signal.
\end{abstract}

\begin{document}

\maketitle

\section{Introduction}

Long-horizon reasoning has become a defining capability of modern LLMs,
driving recent progress in deep research, agentic coding, and multi-step
planning~\citep{mialon2023gaia,liu2023agentbench,wang2025odysseybench,xu2025herobench}. Long-horizon problems in the real world, such as planning a
trip or writing a research report, often span across multiple domains and require different skills at each step. For example, a task may first require math derivation, then use the derived result to plan a schedule, and finally use that schedule to guide information extraction.
This demands that the model fluidly \emph{switch from one
reasoning skill to another} while keeping a single chain of reasoning.
Previous work observed that while frontier models often perform well on single-skill benchmarks~\citep{cobbe2021gsm8k,hendrycks2021math,jain2024livecodebench,lin2024zebralogic,wang2024mmlupro,suzgun2022bbh}, they remain visibly brittle on long-horizon and compositional tasks~\citep{mialon2023gaia,liu2023agentbench,wang2025odysseybench,xu2025herobench,ye2025longproc,kazemi2025bbeh,alazraki2025agentcoma}.
This gap persists even when each component skill is one the model handles well in isolation~\citep{yu2023skillmix,zhao2024skillcomposition,yuan2025rlcompose}, suggesting that handling skill switches is an orthogonal capability to domain-specific competence.

\begin{figure}[t]
    \centering
    \includegraphics[width=\linewidth]{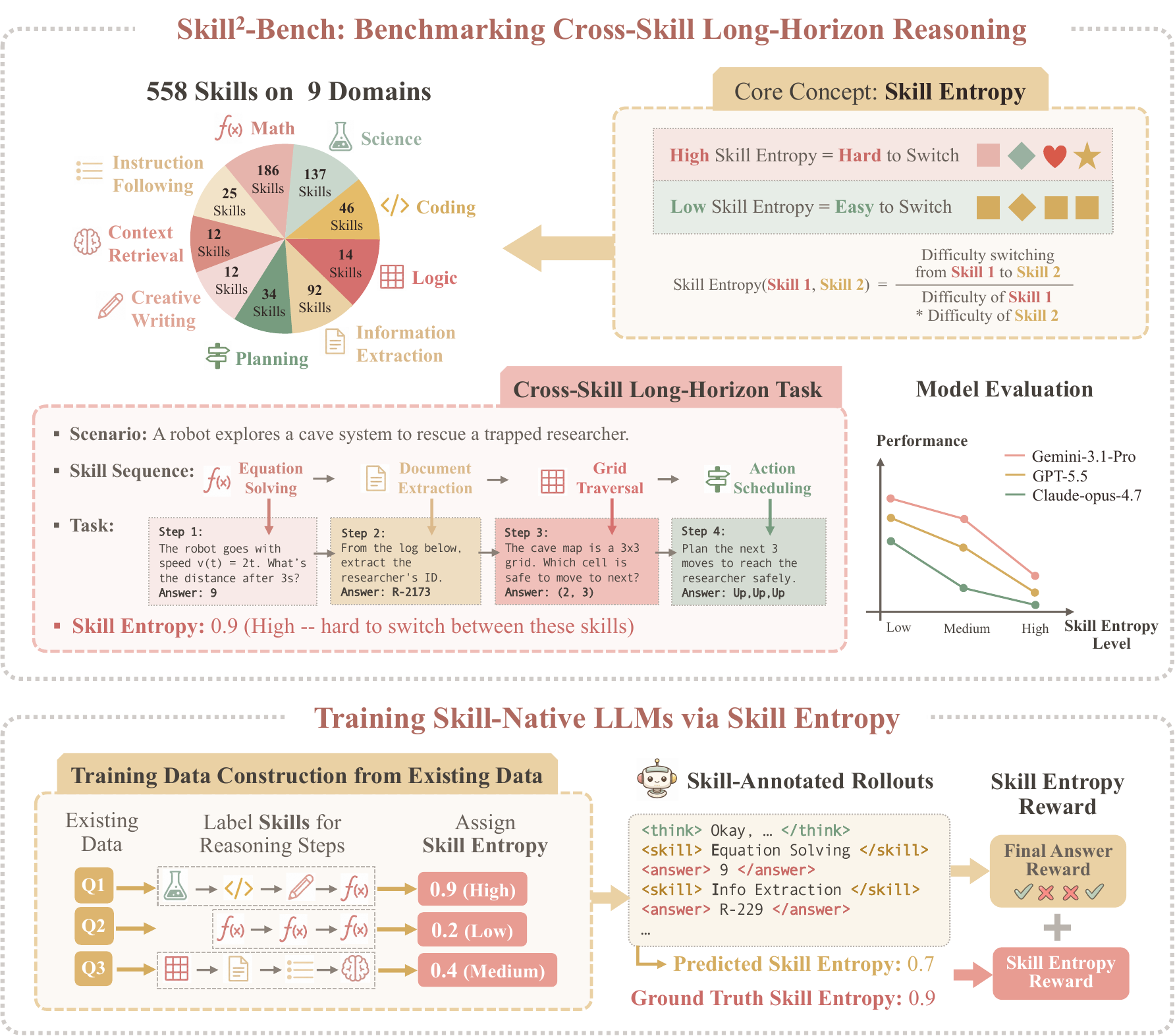}
    \caption{\textbf{Skill Entropy for
Benchmarking and Training Skill-Native LLMs.} We define a \emph{cross-skill long-horizon task} as a sequence of question steps that each invoke a different reasoning skill, and \emph{skill entropy} as a quantity measuring the difficulty of switching from one skill to another. We present: \textbf{(1) \bench{}:} a benchmark over $558$ skills across $9$ verifiable and open-ended domains. Each task is graded by a task-level skill entropy; frontier models degrade nearly monotonically as task-level skill entropy increases. \textbf{(2) Skill-Entropy RL:} an RL framework where the model predicts not only the answer at each step but also the skill used to produce it. The reward combines step-level correctness with a \emph{skill-entropy reward} that measures alignment between the model-predicted skill sequence and the gold skill chain. See \Cref{sec:benchmark} and \Cref{sec:skerl} for details.}
    \label{fig:main-figure}
    \vspace{-10pt}
\end{figure}

To study this skill-switching capability, we formalize a \emph{Cross-Skill Long-Horizon Task} as
a sequence of question steps in which each step invokes a different
reasoning skill, and depends on the answers produced by earlier steps.
Their difficulty is determined not only by which skills they require, but also
by which skill switches their reasoning chain demand. 
Because each skill has its own reasoning style, nearby skills such as math and coding that share symbolic structure would transfer more naturally than distant ones such as creative writing and math derivation~\citep{chen2025symbolicmoe,alazraki2025agentcoma, cheng2025guru}. In general, combining skills within one chain of reasoning is a demanding ability that does not emerge from strong per-skill training alone~\citep{yu2023skillmix,zhao2024skillcomposition,yuan2025rlcompose,wei2026steps}.
To identify where models fail on these tasks, we need a task-level measure that scores how hard a task's skill switches will be. However, existing evaluations~\citep{yu2023skillmix,alazraki2025agentcoma,mialon2023gaia,xu2025herobench} did not provide such a measure.

We close this gap by introducing the concept of \textbf{\emph{Skill Entropy}}
(\Cref{sec:skill_entropy}), a directed pairwise quantity measuring the difficulty of switching from one skill to another (\Cref{eq:pairwise_entropy}).
Aggregating the pairwise skill entropy along a task's skill
sequence gives a scalar task-level skill entropy that quantifies the skill-switching difficulty of each task. Based on this measure, we propose \textbf{\bench{}}
(\Cref{sec:benchmark}), a benchmark of cross-skill long-horizon tasks built
over $558$ skills across 9 verifiable and open-ended domains (math,
coding, science, planning, logic, information extraction, instruction
following, context retrieval, and creative writing), with each task scored
by its task-level skill entropy and sorted into low, medium, and high
difficulty levels.

Evaluating 8 frontier and 4 open-source models on \bench{} surfaces
a structural skill-switching gap (\Cref{sec:evaluation}): accuracy
decreases nearly monotonically as task-level skill entropy rises, and drops by
$-4\sim -13\%$ when the same skill is exercised inside a
cross-skill task rather than a single-skill question. The gap persists even for skills
the model handles well. We trace the dominant failure
mode: at later steps, models tend to reuse the previous step's skill
and answer modality instead of switching to the skill required by the current step.

Motivated by this finding, we turn skill entropy from a benchmark scale
into a training signal. We propose \emph{Skill-Entropy RL}
(\Cref{sec:skerl}), an RL framework where the model commits to a skill
label before each step's answer and is optimized under a reward that
combines step-level correctness with a \emph{skill-entropy reward}
grading the predicted skill chain against the gold one. On \qwen{} and \Sqwen{},
Skill-Entropy RL improves the \bench{} score from $34.4\%$ to $68.4\%$ and from $14.6\%$ to $40.1\%$ respectively. The same pipeline plugs into off-the-shelf
training data such as OpenR1-Math~\citep{openr1}, showing that skill
entropy is a reusable training signal. We summarize our main contributions as follows:
\vspace{-5pt}
\setlength{\leftmargini}{1.2em}
\begin{itemize}
\setlength{\itemsep}{2pt}\setlength{\parskip}{0pt}\setlength{\topsep}{2pt}
\item We formalize \emph{cross-skill long-horizon tasks} and introduce
      \emph{skill entropy}, a directed pairwise measure of how hard it is
      to switch between skills inside a task
      (\Cref{sec:cross_skill_tasks,sec:skill_entropy}).
\item We release \bench{}, a skill-entropy-calibrated benchmark over
      $558$ skills across nine domains. Evaluating 8 frontier and 4 open-source models exposes a skill-switching gap invisible to
      single-domain evaluations (\Cref{sec:evaluation}).
\item We propose \emph{Skill-Entropy RL}, which incorporates the skill entropy reward. It significantly improves \qwen{} and \Sqwen{} on \bench{} score,
      transferring to unseen domains and surpassing strong baselines. It also plugs easily into off-the-shelf training data.
\end{itemize}

\section{Preliminaries}
\label{sec:preliminaries}


\label{sec:cross_skill_tasks}

\textbf{Skills.}
We work with a finite set of \emph{domains}
$\mathcal{D} = \mathcal{D}_{\text{ver}} \cup \mathcal{D}_{\text{open}}$
that splits into \emph{verifiable} and \emph{open-ended} groups.
Each domain $d \in \mathcal{D}$ is backed by a \emph{seed dataset}
$\mathcal{X}_d$ of question--answer pairs $(q, a)$: for
$d \in \mathcal{D}_{\text{ver}}$, $a$ is a ground-truth answer, and for
$d \in \mathcal{D}_{\text{open}}$, $a$ is a grading rubric. Within each
domain $d$, problems use a set of \textbf{\emph{skills}}
$\mathcal{S}_d$ (e.g., {symbolic integration} in {math}, or
{constraint propagation} in {logic}). We write
$\mathcal{S} = \bigcup_{d \in \mathcal{D}} \mathcal{S}_d$ for the full set
of skills, and denote by $d_s \in \mathcal{D}$ the \emph{source domain} of
a skill $s \in \mathcal{S}$, i.e.\ the unique domain in which $s$ was first
seen. Every $(q, a) \in \mathcal{X}_d$ is labeled with a skill
$s \in \mathcal{S}_d$.

\textbf{Cross-Skill Long-Horizon Task.}
A \emph{cross-skill task} is a length-$L$ sequence of question--answer pairs
\begin{equation}
  \tau \;=\; \bigl((q_1, a_1),\, (q_2, a_2),\, \ldots,\, (q_L, a_L)\bigr),
  \qquad L\in \left[2,10\right],
  \label{eq:task_qa}
\end{equation}
paired with a \emph{skill sequence}
\begin{equation}
  \mu(\tau) \;=\; (s_1,\, s_2,\, \ldots,\, s_L),
  \label{eq:task_skills}
\end{equation}
where each step $(q_i, a_i)$ is adapted from the seed dataset
$\mathcal{X}_{d_{s_i}}$ and uses the skill $s_i \in \mathcal{S}$, with
consecutive steps drawn from different domains
($d_{s_i} \neq d_{s_{i+1}}$). The task is tied together by a unifying
\emph{scenario} $\sigma$: an LLM proposer rewrites each $q_i$ to fit
$\sigma$ and to depend on the preceding answer, while keeping the
underlying logic and ground-truth answer of the original seed question.

A model produces a response $\hat{a}_i$ for each step. Each domain $d$ comes
with a domain-specific scorer
$\operatorname{eval}_d : (\hat{a}, a, q) \mapsto [0, 1]$ (e.g., symbolic
equivalence for \text{math}, sandboxed unit-test execution for
\text{coding}; full list in \Cref{app:datasets}). The score of a task is the
average of its per-step scores.

\section{\bench{}: Benchmarking Cross-Skill Long-Horizon Reasoning}
\label{sec:benchmark}

\begin{table}[t]
\centering
\setlength{\tabcolsep}{2.3pt}
\renewcommand{\arraystretch}{1.18}
\resizebox{\textwidth}{!}{%
\begin{tabular}{>{\normalsize}l@{\hspace{-10pt}} l@{\hspace{-10pt}} >{\large}c@{\hspace{-2.5pt}}>{\normalsize} c@{\hspace{-1.5pt}} >{\small}l}
\toprule
\textbf{Domain} & \textbf{Seed Dataset} & \multicolumn{1}{c}{\normalsize \textbf{Verifiable}} & \multicolumn{1}{c}{\normalsize \textbf{\# Skills}} & \multicolumn{1}{c}{\normalsize \textbf{Example Skills}} \\
\midrule
\raisebox{-2.5pt}{\includegraphics[width=14pt]{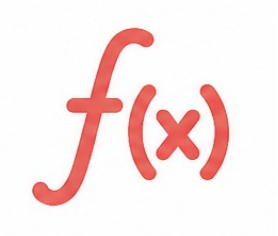}}\hspace{-0.5pt} Math                    & OpenR1-Math~\citep{openr1}                       & \textcolor{green!80!black}{\textbf{\checkmark}} & 186 & Geometric Proofs, Trigonometric Functions \\
\raisebox{-2pt}{\includegraphics[width=12pt]{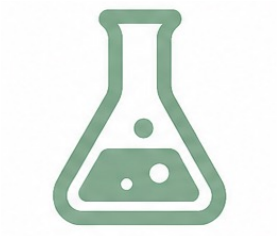}}\hspace{1.1pt} Science                 & MMLU-Pro~\citep{wang2024mmlupro}                          & \textcolor{green!80!black}{\textbf{\checkmark}} & 137 & Statistical Modeling, Financial Mathematics \\
\raisebox{-1pt}{\includegraphics[width=12pt]{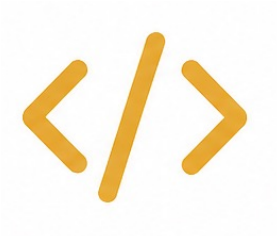}}\hspace{3pt}Coding                  & LiveCodeBench~\citep{jain2024livecodebench}                & \textcolor{green!80!black}{\textbf{\checkmark}} & 46 & Graph Theory, String Manipulation \\
\multirow{1}{*}{\raisebox{-1.5pt}{\includegraphics[width=12pt]{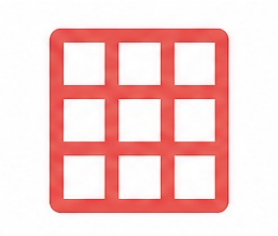}}\hspace{3pt}Logic}                   & ZebraLogicBench~\citep{lin2024zebralogic},      & \multirow{1}{*}{\textcolor{green!80!black}{\textbf{\checkmark}}} & \multirow{1}{*}{14} & \multirow{1}{*}{Constraint Satisfaction, Deductive Reasoning} \\
                        & Guru-RL-92k~\citep{cheng2025guru}                & & & \\
\multirow{1}{*}{\raisebox{-2pt}{\includegraphics[width=13pt]{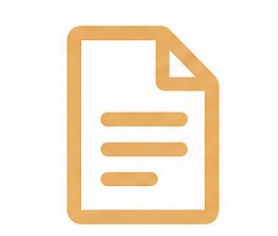}}\hspace{3pt}Information}  & WikiTable~\citep{pasupat2015wikitable},   & \multirow{1}{*}{\textcolor{green!80!black}{\textbf{\checkmark}}} & \multirow{1}{*}{92} & \multirow{1}{*}{Table Data Analysis, Information Retrieval} \\
                         
\hspace{16.5pt}Extraction & WebSRC~\citep{chen2021websrc}                & & & \\
\raisebox{-1.5pt}{\includegraphics[width=13pt]{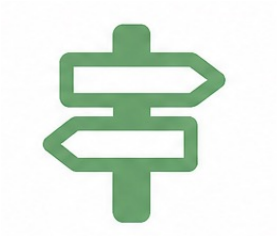}}\hspace{3pt}Planning                & NaturalPlan~\citep{zheng2024naturalplan}                       & \textcolor{green!80!black}{\textbf{\checkmark}} & 34 & Activity Scheduling, Travel Itinerary Planning \\
\raisebox{-2pt}{\includegraphics[width=12pt]{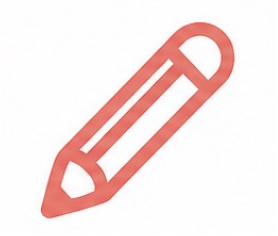}}\hspace{3pt}Creative Writing        & \multicolumn{1}{c}{---}          & \textcolor{pink}{$\boldsymbol{\times}$} & 12 & Brainstorming, Storytelling \\
\raisebox{-2pt}{\includegraphics[width=12pt]{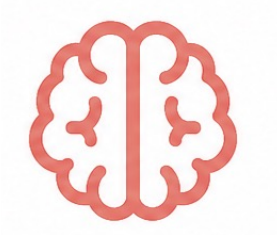}}\hspace{3pt}Context Retrieval        & \multicolumn{1}{c}{---}          & \textcolor{pink}{$\boldsymbol{\times}$} & 12 & Context Recall, Entity Tracking \\
\raisebox{-2pt}{\includegraphics[width=12pt]{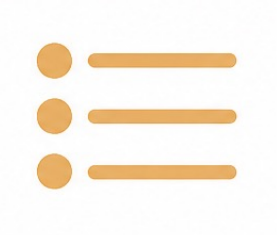}}\hspace{3pt}Instruction Following   & \multicolumn{1}{c}{---}           & \textcolor{pink}{$\boldsymbol{\times}$} & 25 & Include Keywords, JSON Format \\
\bottomrule
\end{tabular}}
\vspace{4pt}
\caption{\textbf{\bench{} includes 9 Domains with 558 labeled skills in total.} Verifiable domains use question--answer pairs adapted from their seed datasets; open-ended domains use LLM-generated question--rubric pairs scored by LLM judges
(\Cref{app:datasets}). The full per-domain skill list is in
\Cref{app:skill_bank}.}
\label{tab:domain-overview}
\vspace{-10pt}
\end{table}

We introduce \textbf{\bench{}}, a benchmark that evaluates how reliably LLMs switch between skills inside a long-horizon task. The difficulty of such a task is not visible from its surface form: two tasks with the same length and skills can still differ greatly in how hard their skill switches are. To make this difficulty measurable, we first develop the \emph{skill entropy} framework (\Cref{sec:skill_entropy}), a scalar score that quantifies how hard the skill switches in a task are. Skill entropy then guides the construction of cross-skill long-horizon tasks at controlled difficulty levels (\Cref{sec:synthesis}), which we use to evaluate how skill-switching performance varies across models and difficulty levels (\Cref{sec:evaluation}).

\subsection{Skill Entropy}
\label{sec:skill_entropy}

We introduce \textbf{\emph{Skill Entropy}}, $\text{SkE}(s_a, s_b)$,
a directional pairwise quantity on two skills $s_a, s_b \in \mathcal{S}$.
It measures how hard it is to switch \emph{from} $s_a$ \emph{to} $s_b$
inside one cross-skill task. We use the term ``entropy'' by analogy, suggesting that harder skill switches make the next correct response less predictable under a fixed reference model. A high skill entropy means accuracy
\emph{drops} sharply when the model answers a question on $s_b$ on top of
an answer it produced with $s_a$, compared to handling each skill in
isolation. The order matters: $\text{SkE}(s_a, s_b)$ and
$\text{SkE}(s_b, s_a)$ are computed on different two-step pairs and in
general differ.
Operationally, $\text{SkE}$ is a smoothed ratio between average
single-skill accuracy and cross-skill accuracy under a fixed reference
model, with $\text{SkE}\approx 1$ indicating that chaining the two skills adds little difficulty over the per-skill baselines.

\textbf{Pairwise skill entropy.}
We fix a reference model so that SkE is held constant across evaluated models and can be used as a common difficulty scale. Under the reference
model, let $\text{Accuracy}(s)$ be the accuracy on multi-step
questions using only skill $s$ in domain $d_s$,
and $\text{Accuracy}(s_a, s_b)$ the average per-step accuracy on
two-step cross-skill pairs whose first step uses $s_a$ and second step uses $s_b$. 
With Laplace smoothing $\alpha = 0.1$, we define
{\small
  \begin{equation}
  \text{SkE}(s_a, s_b)
  \;=\; \frac{\tfrac{1}{2}\bigl(\text{Accuracy}(s_a)+\text{Accuracy}(s_{b})\bigr) + \alpha}
              {\text{Accuracy}(s_a, s_b) + \alpha}
  \quad
  \begin{cases}
    >1, & \text{hard to switch from }s_a\text{ to }s_b,\\
    \leq 1, & \text{easy to switch from }s_a\text{ to }s_b.
  \end{cases}
  \label{eq:pairwise_entropy}
\end{equation}}

In practice, we
approximate $\text{SkE}(s_a, s_b)$ by evaluating skill-to-domain switches, as detailed in \Cref{app:ske_factorization}.

\textbf{Task-level skill entropy.}
The skill entropy of a cross-skill task $\tau$ with skill sequence
$\mu(\tau) = (s_1, \ldots, s_L)$ is the average pairwise skill entropy
along the task's directed skill switches $s_i \to s_{i+1}$,
\begin{equation}
  \text{SkE}(\tau)
  \;=\;
  \frac{1}{L-1}\sum_{i=1}^{L-1} \text{SkE}(s_i, s_{i+1}).
  \label{eq:task_entropy}
\end{equation}
We then split tasks into \emph{low}, \emph{medium}, and \emph{high}
skill-entropy levels by thresholding $\text{SkE}(\tau)$ at two boundaries
$(\theta_\ell, \theta_h)$ taken from the empirical distribution of pairwise
scores. This split examines whether models that score similarly on
low-skill-entropy tasks pull apart as the cross-skill demands grow.

\begin{figure}
    \centering
    \includegraphics[width=0.54\linewidth]{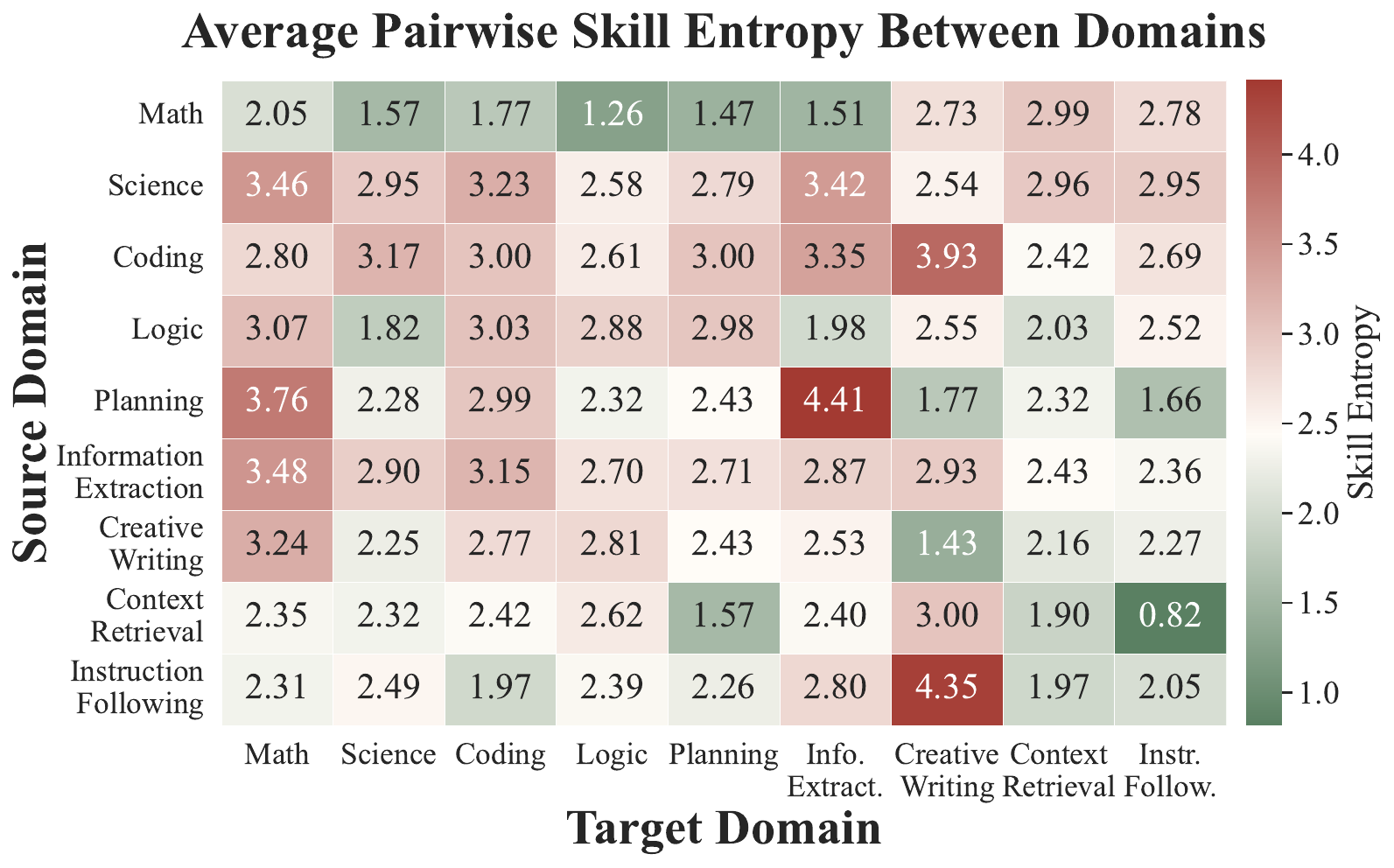} \hspace{3pt}
    \includegraphics[width=0.44\linewidth]{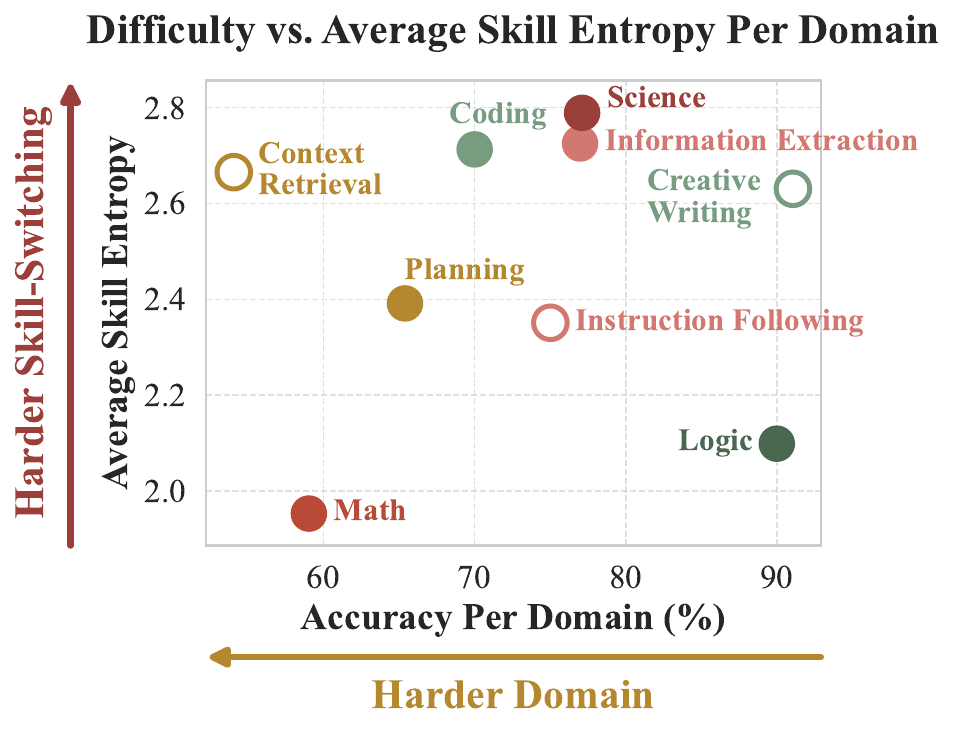}
    \caption{\textbf{Visualization of skill entropy across 9 \bench{} domains.}
    \textbf{Left:} pairwise skill entropy averaged by (source, target) domain pair. Intuitively, \textcolor{red}{\text{larger values}} in a cell mean that switching skills from the source domain to the target domain is \textcolor{red}{\text{harder}}.
    \textbf{Right:} per-domain accuracy vs. per-domain skill entropy. Lower accuracy indicates a harder domain; higher skill entropy indicates that switching into or out of that domain is harder.
    }
    \label{fig:skill_entropy_overview}
    \vspace{-5pt}
\end{figure}

\subsection{Construction of Cross-Skill Long-Horizon Tasks}
\label{sec:synthesis}

We construct cross-skill long-horizon tasks in three stages: (1) build a per-domain
\emph{skill bank}, the set of skill labels annotated on the seed questions of that domain, from seed datasets;
(2) derive pairwise skill entropies by evaluating a reference model on single-skill vs. cross-skill questions;
(3) sample a skill sequence at a target skill-entropy level, then rewrite the corresponding seed questions into a single coherent scenario with an LLM \emph{proposer} and filter the resulting tasks with a \emph{verifier}. Full prompts, the skill bank, a worked example, and the
verifier checklist are in \Cref{app:synthesis}.

\textbf{(1) Constructing skill bank from seed datasets:}
For each verifiable domain, we prompt an LLM to label every seed question with $3$--$5$ fine-grained skills, then cluster the resulting labels via embedding, and manually review each cluster to keep skill granularity comparable across domains. For the three open-ended domains, we prompt an LLM to generate a skill list directly for each domain. \Cref{tab:domain-overview} summarizes the 9 domains and resulting skills; full per-domain procedures, splits, and licenses are in \Cref{app:datasets,app:skill_bank}.

\textbf{(2) Skill entropy derivation via evaluating reference model:}
With Claude-opus-4.7 as the reference model, we run it on single-skill seed questions to measure its baseline accuracy on each skill and each domain in isolation, and on two-step cross-skill tasks to measure how that accuracy changes when we chain questions together. Plugging the resulting accuracies into \Cref{eq:pairwise_entropy} with smoothing constant $0.1$ gives the pairwise skill entropy. Sampling counts, the multi-step protocol, and the choice of reference model are in \Cref{app:entropy_estimation,app:reference_model}.

\textbf{(3) Skill sequence sampling \& task scenario proposal.}
For each task we sample a length between $2$ and $10$ and a skill sequence at one of three target skill-entropy levels (low / medium / high, see \Cref{eq:task_entropy}). For skills from verifiable domains, we draw one seed question--answer pair per skill and prompt an LLM \emph{proposer} to fuse them into a single coherent scenario in which each step depends on the preceding one; for skills from the three open-ended domains, the LLM directly generates question--rubric pairs under the same scenario. A separate \emph{verifier} filters out any task that fails its checklist. The sequence-sampling procedure, proposer prompt, open-ended generation prompt, and verifier checklist are in \Cref{app:sequence_sampling,app:proposer,app:example,app:filtering}.

\textbf{The Landscape of Skill Entropy Across Domains.}
\Cref{fig:skill_entropy_overview} visualizes the resulting skill entropies. The pairwise skill-entropy (left) varies substantially across (source, target) domain pairs, with planning to information extraction being the hardest skill switch.
The right panel shows that per-domain skill entropy is largely decoupled from domain difficulty itself. Science has a high domain accuracy (easy domain) yet the highest skill entropies (hard skill-switching), suggesting that science skills are easier but highly domain-specific. Therefore, skill entropy exposes another dimension of task difficulty: how well a model handles a domain in isolation does not predict how hard that domain is to switch from/to others.

\begin{table}[t]
\centering
\small
\setlength{\tabcolsep}{3.5pt}
\renewcommand{\arraystretch}{1.12}
\resizebox{\textwidth}{!}{%
\begin{tabular}{l l@{\hspace{1.8pt}} ccccc c >{\color{red}\bfseries\boldmath}l cccc}
\toprule
\multirow{2}{*}{\textbf{Model}}
& \multicolumn{7}{c}{\textbf{Domain Accuracy}}
&
& \multicolumn{4}{c}{\textbf{\bench{} Performance}} \\
\cmidrule(lr){2-9} \cmidrule(lr){10-13}
& Setting & \textbf{Coding} & \textbf{Logic} & \textbf{Math} & \textbf{Planning} & \textbf{Science} & \textbf{Avg.} & $\mathbf{\Delta}$ \textbf{(\%)}
& \textbf{Low} & \textbf{Medium} & \textbf{High} & \textbf{Avg.}\\
\midrule
\rowcolor{red!18}\multicolumn{13}{l}{\textit{\text{\# Frontier models}}} 
\vspace{2pt}\\
\multirow{2}{*}{\textbf{Claude-haiku-4.5}}
 & \single{} & 59.0 & 62.9 & 41.9 & \phantom{0}5.1 & 74.9 & 48.8
 & \multirow{2}{*}{$-$7.8}
 & \multirow{2}{*}{\shortstack{62.3 \\ {\tiny $\pm 1.6$}}} & \multirow{2}{*}{\shortstack{56.3 \\ {\tiny $\pm 1.2$}}} & \multirow{2}{*}{\shortstack{54.9 \\ {\tiny $\pm 1.4$}}} & \multirow{2}{*}{\shortstack{57.8 \\ {\tiny $\pm 1.0$}}} \\
 & \cc\cross{} & \cc 58.0 & \cc 40.0 & \cc 36.7 & \cc \phantom{0}2.8 & \cc 67.2 & \cc 40.9 & & & & & \\
\addlinespace[2pt]
\multirow{2}{*}{\textbf{Claude-sonnet-4.5}}
 & \single{} & 55.7 & 47.4 & 45.3 & 43.5 & 68.5 & 52.1
 & \multirow{2}{*}{$-$6.5}
 & \multirow{2}{*}{\shortstack{67.2 \\ {\tiny $\pm 1.0$}}} & \multirow{2}{*}{\shortstack{62.0 \\ {\tiny $\pm 1.4$}}} & \multirow{2}{*}{\shortstack{59.5 \\ {\tiny $\pm 1.8$}}} & \multirow{2}{*}{\shortstack{62.9 \\ {\tiny $\pm 0.8$}}} \\
 & \cc\cross{} & \cc 48.0 & \cc 58.6 & \cc 34.4 & \cc 20.6 & \cc 66.5 & \cc 45.6 & & & & & \\
\addlinespace[2pt]
\multirow{2}{*}{\textbf{Claude-opus-4.7}}
 & \single{} & 66.8 & 95.0 & 62.3 & 67.3 & 83.4 & 75.0
 & \multirow{2}{*}{$-$4.3}
 & \multirow{2}{*}{\shortstack{72.2 \\ {\tiny $\pm 0.8$}}} & \multirow{2}{*}{\shortstack{69.2 \\ {\tiny $\pm 1.0$}}} & \multirow{2}{*}{\shortstack{68.6 \\ {\tiny $\pm 1.2$}}} & \multirow{2}{*}{\shortstack{70.0 \\ {\tiny $\pm 0.6$}}} \\
 & \cc\cross{} & \cc 64.8 & \cc 88.8 & \cc 61.8 & \cc 56.0 & \cc 82.1 & \cc 70.7 & & & & & \\
\addlinespace[2pt]
\multirow{2}{*}{\textbf{Gemini-3.1-flash}}
 & \single{} & 53.7 & 52.9 & 49.2 & 26.5 & 79.7 & 52.4
 & \multirow{2}{*}{$-$9.1}
 & \multirow{2}{*}{\shortstack{65.9 \\ {\tiny $\pm 1.4$}}} & \multirow{2}{*}{\shortstack{60.7 \\ {\tiny $\pm 1.6$}}} & \multirow{2}{*}{\shortstack{60.3 \\ {\tiny $\pm 1.2$}}} & \multirow{2}{*}{\shortstack{62.3 \\ {\tiny $\pm 1.0$}}} \\
 & \cc\cross{} & \cc 38.3 & \cc 41.0 & \cc 45.9 & \cc 16.3 & \cc 75.0 & \cc 43.3 & & & & & \\
\addlinespace[2pt]
\multirow{2}{*}{\textbf{Gemini-3.1-pro}}
 & \single{} & 66.0 & 97.1 & 72.9 & 88.4 & 84.1 & 81.7
 & \multirow{2}{*}{$-$5.4}
 & \multirow{2}{*}{\shortstack{77.1 \\ {\tiny $\pm 0.6$}}} & \multirow{2}{*}{\shortstack{75.2 \\ {\tiny $\pm 0.8$}}} & \multirow{2}{*}{\shortstack{72.2 \\ {\tiny $\pm 1.0$}}} & \multirow{2}{*}{\shortstack{74.8 \\ {\tiny $\pm 0.6$}}} \\
 & \cc\cross{} & \cc 66.5 & \cc 94.3 & \cc 70.7 & \cc 76.5 & \cc 73.6 & \cc 76.3 & & & & & \\
\addlinespace[2pt]
\multirow{2}{*}{\textbf{GPT-5.4-mini}}
 & \single{} & 50.8 & 18.6 & 14.3 & 17.7 & 47.9 & 29.9
 & \multirow{2}{*}{$-$10.0}
 & \multirow{2}{*}{\shortstack{42.6 \\ {\tiny $\pm 2.2$}}} & \multirow{2}{*}{\shortstack{39.2 \\ {\tiny $\pm 2.0$}}} & \multirow{2}{*}{\shortstack{42.7 \\ {\tiny $\pm 1.8$}}} & \multirow{2}{*}{\shortstack{41.5 \\ {\tiny $\pm 1.4$}}} \\
 & \cc\cross{} & \cc 28.2 & \cc \phantom{0}3.2 & \cc 11.5 & \cc 12.4 & \cc 44.0 & \cc 19.9 & & & & & \\
\addlinespace[2pt]
\multirow{2}{*}{\textbf{O4-mini}}
 & \single{} & 60.0 & 94.3 & 58.7 & 32.5 & 50.0 & 59.1
 & \multirow{2}{*}{$-$6.9}
 & \multirow{2}{*}{\shortstack{59.9 \\ {\tiny $\pm 1.2$}}} & \multirow{2}{*}{\shortstack{60.8 \\ {\tiny $\pm 1.0$}}} & \multirow{2}{*}{\shortstack{58.6 \\ {\tiny $\pm 1.6$}}} & \multirow{2}{*}{\shortstack{59.8 \\ {\tiny $\pm 0.8$}}} \\
 & \cc\cross{} & \cc 62.5 & \cc 82.9 & \cc 50.4 & \cc \phantom{0}5.7 & \cc 59.5 & \cc 52.2 & & & & & \\
\addlinespace[2pt]
\multirow{2}{*}{\textbf{GPT-5.5}}
 & \single{} & 68.1 & 97.3 & 67.1 & 58.2 & 73.2 & 72.8
 & \multirow{2}{*}{$-$4.6}
 & \multirow{2}{*}{\shortstack{74.6 \\ {\tiny $\pm 0.8$}}} & \multirow{2}{*}{\shortstack{71.5 \\ {\tiny $\pm 1.0$}}} & \multirow{2}{*}{\shortstack{71.9 \\ {\tiny $\pm 0.8$}}} & \multirow{2}{*}{\shortstack{72.7 \\ {\tiny $\pm 0.6$}}} \\
 & \cc\cross{} & \cc 68.5 & \cc 95.3 & \cc 64.8 & \cc 44.5 & \cc 67.8 & \cc 68.2 & & & & & \\
\midrule
\rowcolor{yellow!25}\multicolumn{13}{l}{\textit{\text{\# Open-source models}}} \vspace{2pt}\\
\multirow{2}{*}{\textbf{Olmo-3-7B-Think}}
 & \single{} & 12.6 & \phantom{0}2.1 & 27.6 & \phantom{0}2.1 & 27.5 & 14.4
 & \multirow{2}{*}{$-$7.2}
 & \multirow{2}{*}{\shortstack{12.8 \\ {\tiny $\pm 2.8$}}} & \multirow{2}{*}{\shortstack{12.8 \\ {\tiny $\pm 2.4$}}} & \multirow{2}{*}{\shortstack{\phantom{0}7.4 \\ {\tiny $\pm 2.0$}}} & \multirow{2}{*}{\shortstack{11.0 \\ {\tiny $\pm 1.8$}}} \\
 & \cc\cross{} & \cc \phantom{0}3.0 & \cc \phantom{0}1.3 & \cc 13.9 & \cc \phantom{0}2.2 & \cc 15.6 & \cc \phantom{0}7.2 & & & & & \\
\addlinespace[2pt]
\multirow{2}{*}{\textbf{Qwen3-4B}}
 & \single{} & 30.5 & 18.6 & 25.7 & 20.2 & 47.2 & 28.4
 & \multirow{2}{*}{$-$12.6}
 & \multirow{2}{*}{\shortstack{32.4 \\ {\tiny $\pm 2.4$}}} & \multirow{2}{*}{\shortstack{24.9 \\ {\tiny $\pm 2.0$}}} & \multirow{2}{*}{\shortstack{26.0 \\ {\tiny $\pm 1.6$}}} & \multirow{2}{*}{\shortstack{27.8 \\ {\tiny $\pm 1.4$}}} \\
 & \cc\cross{} & \cc 25.5 & \cc \phantom{0}7.1 & \cc 12.0 & \cc \phantom{0}5.3 & \cc 29.2 & \cc 15.8 & & & & & \\
\addlinespace[2pt]
\multirow{2}{*}{\textbf{Qwen3-8B}}
 & \single{} & 49.0 & 10.0 & 37.0 & 13.8 & 47.2 & 31.4
 & \multirow{2}{*}{$-$9.3}
 & \multirow{2}{*}{\shortstack{39.1 \\ {\tiny $\pm 1.8$}}} & \multirow{2}{*}{\shortstack{33.2 \\ {\tiny $\pm 2.2$}}} & \multirow{2}{*}{\shortstack{31.5 \\ {\tiny $\pm 2.0$}}} & \multirow{2}{*}{\shortstack{34.6 \\ {\tiny $\pm 1.2$}}} \\
 & \cc\cross{} & \cc 37.4 & \cc 12.9 & \cc 12.6 & \cc 13.3 & \cc 34.2 & \cc 22.1 & & & & & \\
\addlinespace[2pt]
\multirow{2}{*}{\textbf{Qwen3-32B}}
 & \single{} & 50.1 & 13.0 & 29.1 & 22.2 & 54.1 & 33.7
 & \multirow{2}{*}{$-$7.7}
 & \multirow{2}{*}{\shortstack{46.0 \\ {\tiny $\pm 1.4$}}} & \multirow{2}{*}{\shortstack{41.0 \\ {\tiny $\pm 1.6$}}} & \multirow{2}{*}{\shortstack{41.0 \\ {\tiny $\pm 1.2$}}} & \multirow{2}{*}{\shortstack{42.7 \\ {\tiny $\pm 1.0$}}} \\
 & \cc\cross{} & \cc 44.9 & \cc 13.6 & \cc 14.8 & \cc 14.1 & \cc 42.6 & \cc 26.0 & & & & & \\
\bottomrule
\end{tabular}}
\vspace{0pt}
\caption{\textbf{Evaluation results on \bench{}.} We present two statistics: \textbf{(1) Domain Accuracy,} reported in \emph{single-skill (``\single{}'')} setting where model answers a single-skill multi-step question, and \emph{cross-skill (``\cross{}'')}  setting where each evaluated questions are put into a long-horizon cross-skill task. Domain accuracy consistently drops from single-skill to cross-skill setting, especially for less capable models.
\textbf{(2) \bench{} Performance,}
reported in three skill entropy level low / medium / high. We observe a broadly monotonic accuracy drop from low to high skill entropy across nearly all
models, confirming that skill entropy tracks cross-skill task difficulty.
}
\label{tab:skebench-results}
\vspace{-5pt}
\end{table}

\subsection{Evaluating LLMs' Cross-Skill Capabilities via Skill Entropy}
\label{sec:evaluation}

\textbf{Experimental setup.}
\label{par:eval_protocol}
\textbf{(1) Models:} Eight frontier models (Claude-haiku-4.5,
Claude-sonnet-4.5, Claude-opus-4.7, Gemini-3.1-flash, Gemini-3.1-pro,
GPT-5.4-mini, O4-mini, GPT-5.5) and four open-source models
(Qwen3-4B, Qwen3-8B, Qwen3-32B, Olmo-3-7B-Think). 
\textbf{(2) Tasks:} A heldout test set of $300$ cross-skill tasks synthesized via
\Cref{sec:synthesis}, balanced across the three skill-entropy levels
and the nine domains, with length 2 to 10.
\textbf{(3) Evaluation Settings:} We query each model in two modes.
The single-skill mode asks every step in isolation as a multi-step
question using only one skill, while the cross-skill
mode inputs the full long-horizon task and asks the model to answer
every step in order. We sample 4 times at temperature $0.7$ (see \Cref{app:eval_settings}).
\textbf{(4) Metrics:} Verifiable steps are graded by per-step correctness, and open-ended steps use an LLM judge (Claude-opus-4.7) in
$[0, 1]$ against a given rubric. \bench{} performance is calculated as model's average per-step scores. \Cref{tab:skebench-results} reports per-domain accuracy on the five most populated verifiable domains (coding, logic, math, planning, science); per-domain results for the remaining four (information extraction, Creative Writing, Context Retrieval, instruction following) are in \Cref{app:full_results}.

\textbf{Performance drops as skill entropy increases.}
\Cref{tab:skebench-results} shows that the \bench{} score decreases as
task-level skill entropy moves from low to high for nearly all models.
The largest frontier models (Claude-opus-4.7, GPT-5.5, Gemini-3.1-pro)
drop by only a few points, while smaller frontier and open-source
models (Claude-haiku-4.5, Qwen3-4B) drop more sharply. 
This consistent trend confirms that skill entropy
tracks cross-skill task difficulty.

\textbf{Cross-skill setting consistently hurts accuracy.}
Every frontier model loses accuracy when the same skill is exercised
inside a cross-skill task, with the per-model drop ranging from
$-4\%$ to $-10\%$, with the largest drops observed on Planning skills. The drop persists
even on skills the model nearly saturates in single-skill mode, such as Logic skills. 
\bench{} thus exposes a
failure mode that single-skill evaluations fail to detect.

\textbf{Failure mode analysis: skill choice \& answer modality in later steps are heavily influenced by previous steps.}
The largest cross-skill failure mode is that, on a step deep into a
task, models tend to carry over the skill and answer modality from
previous steps instead of switching to the ones the current step
requires. \Cref{fig:case-study} shows a representative case: after a
Geometric Calculation step, the base model reuses a math skill with
a short numeric answer on the next step, where Theme Creation with a
long passage is required. \Cref{fig:failure_mode}
(\Cref{app:full_results}) quantifies this failure mode: across the
strongest frontier models, picking a wrong skill at a step
roughly halves its accuracy.
\section{Training Skill-Native LLMs via Skill Entropy}
\label{sec:skerl}

\Cref{sec:evaluation} showed that even strong models lose accuracy as
task-level skill entropy grows, with smaller open-source models suffering
the largest drops. We now ask whether the same skill-entropy signal that
\emph{exposes} this gap can also be used to \emph{train} a model to
close it. We propose a two-stage pipeline for RL with skill entropy that
first teaches the model to emit a structured skill plan together with
its answer, and then optimizes that plan with reinforcement learning
under a reward that grades whether the predicted skill structure
matches the actual difficulty of the task.

\subsection{Skill Entropy as a Training Signal}
\label{sec:skerl_method}

\textbf{Skill-annotated response format.}
Given a cross-skill task $\tau = ((q_1,a_1),\ldots,(q_L,a_L))$ from
\Cref{eq:task_qa}, the model produces a response that pairs each step
with the skill it invokes:
\begin{prompt}[equal height group=pairA,title=Skill-annotated response format]{yellow!10}{yellow!90}
  <think> [Reasoning Trace] </think>

<skill> Domain\_1, Skill\_1 </skill><answer> Step\_1 Answer </answer>

                                ...

<skill> Domain\_L, Skill\_L </skill><answer> Step\_L Answer (Final Answer) </answer>
\end{prompt}
The response parses into a predicted skill sequence
$\hat{\mu}(\tau) = (\hat{s}_1, \ldots, \hat{s}_L)$ and answer sequence
$(\hat{a}_1, \ldots, \hat{a}_L)$; the regex parser and an end-to-end SFT
trace are in \Cref{app:skerl_format}.

\textbf{Reward.}
We train with GRPO under a per-task reward that combines an
\emph{answer reward} and a \emph{skill-entropy reward}, both in $[0,1]$:
\begin{equation}
  r \;=\; \lambda_{\text{ans}}\, r_{\text{ans}}
       \;+\; \lambda_{\text{ent}}\, r_{\text{ent}}.
  \label{eq:skerl-reward}
\end{equation}
$r_{\text{ans}}$ is the mean per-step accuracy of the response under the
per-domain scorers of \Cref{sec:cross_skill_tasks}. $r_{\text{ent}}$
grades the predicted skill plan against the gold plan by their
task-level skill-entropy ranks on the training distribution: writing
$\hat{\rho}, \rho^{\star} \in [0,1]$ for the rank of the predicted and
gold skill entropies (\Cref{eq:task_entropy}),
\begin{equation}
  r_{\text{ent}} \;=\; 1 - \bigl|\,\hat{\rho} - \rho^{\star}\,\bigr|.
  \label{eq:skerl-ent}
\end{equation}
To compute $\hat{\rho}$, we first map each predicted skill via
embedding similarity to its closest entry in the skill bank, then evaluate the skill entropy on these mapped skills. $r_{\text{ent}}$
is thus larger when the predicted plan has a skill sequence closer to the
gold plan, allowing the model to substitute semantically similar skills. Formal reward definitions,
embedding-based skill matching, and GRPO hyperparameters are in
\Cref{app:skerl_reward,app:skerl_skill_match,app:skerl_training}.

\begin{table}[t]
\centering
\small
\setlength{\tabcolsep}{2.2pt}
\renewcommand{\arraystretch}{0.95}
\resizebox{\textwidth}{!}{%
\begin{tabular}{l@{\hspace{5pt}} ccccc cccc c @{\hspace{1.1pt}}c}
\toprule
\multirow{2}{*}{\textbf{Method}}
& \multicolumn{6}{c}{\textbf{Verifiable Domains}} & \multicolumn{3}{c}{\textbf{Open-Ended Domains}}
&
& \multirow{2.1}{*}{\textbf{\bench{}}} \\
\cmidrule(lr){2-7} \cmidrule(lr){8-10}
& {\footnotesize\textbf{Math}} & {\footnotesize\textbf{Coding}} & {\footnotesize\textbf{Science}} & {\footnotesize\textbf{Planning}} & {\footnotesize\textbf{Logic}}
& {\footnotesize\makecell{\textbf{Info.}\\\textbf{Extraction}}}
& {\footnotesize\makecell{\textbf{Instruction}\\\textbf{Following}}}
& {\footnotesize\makecell{\textbf{Ctx.}\\\textbf{Retrieval}}}
& {\footnotesize\makecell{\textbf{Creative}\\\textbf{Writing}}}
& &  \multirow{0.1}{*}{\textbf{Performance}}\\
\midrule
\rowcolor{green!30}\multicolumn{12}{l}{\textit{\text{\# Qwen3-4B-Instruct}}} \vspace{3pt}\\
Base model
 & 14.0 & 26.3 & 34.5 & \phantom{0}5.9 & 12.9 & 47.8 & 64.2 & 59.2 & 68.8 & & 34.4 \\
\addlinespace[2pt]
SFT
 & 37.5 & 33.3 & 61.6 & 55.6 & 38.6 & 69.4 & 69.4 & 55.4 & 60.2 & & 55.8 \\
\addlinespace[2pt]
GRPO
 & 44.7 & 42.8 & 54.6 & \textbf{67.2} & 41.4 & 73.7 & 75.0 & 63.9 & 68.8 & & 58.8 \\
\addlinespace[2pt]
Skill-Distill~\citep{zhang2026skilldistill}
 & 45.4 & 41.5 & 61.5 & 53.5 & 28.7 & 72.4 & 71.7 & 55.3 & 69.1 & & 58.1 \\
\addlinespace[2pt]
SkillRL~\citep{xia2026skillrl}
 & 44.8 & 43.4 & 63.6 & 53.5 & 34.4 & 72.8 & 73.2 & 62.7 & 62.4 & & 59.3 \\
\addlinespace[2pt]
STAT~\citep{he2025stat}
 & 45.7 & 47.0 & 68.1 & 56.5 & 35.9 & 74.9 & \textbf{76.7} & 60.0 & 67.1 & & 61.4 \\
\addlinespace[2pt]
\textbf{Skill-Entropy RL}
 & \textbf{49.3} & \textbf{47.8} & \textbf{71.1} & 55.8 & \textbf{47.1} & \textbf{76.8} & 75.2 & \textbf{64.7} & \textbf{85.6} & & \textbf{68.4} \\
\addlinespace[2pt]
\midrule
\rowcolor{yellow!25}\multicolumn{12}{l}{\textit{\text{\# Qwen3-1.7B}}} \vspace{3pt}\\
Base model
 & 11.7 & 13.0 & 13.3 & \phantom{0}9.4 & \phantom{0}9.4 & 17.7 & 16.5 & 14.1 & 17.8 & & 14.6 \\
\addlinespace[2pt]
SFT
 & 12.3 & 27.5 & 43.7 & 34.8 & 17.1 & 37.8 & 32.8 & 40.6 & 42.9 & & 30.6 \\
\addlinespace[2pt]
GRPO
 & 14.3 & 30.3 & 44.4 & 36.1 & 15.7 & 37.2 & 26.9 & 44.3 & 48.9 & & 32.2 \\
\addlinespace[2pt]
Skill-Distill~\citep{zhang2026skilldistill}
 & 16.5 & 32.0 & 37.2 & 32.9 & 11.6 & 38.6 & \textbf{39.3} & 46.7 & 42.2 & & 32.4 \\
\addlinespace[2pt]
SkillRL~\citep{xia2026skillrl}
 & 14.6 & 30.2 & 39.6 & 35.1 & 14.0 & 43.2 & 35.1 & 44.0 & 47.9 & & 32.7 \\
\addlinespace[2pt]
STAT~\citep{he2025stat}
 & 15.3 & 33.3 & 40.3 & 32.8 & 11.6 & 38.3 & 35.0 & \textbf{52.4} & 48.6 & & 33.0 \\
\addlinespace[2pt]
\textbf{Skill-Entropy RL}
 & \textbf{27.4} & \textbf{36.8} & \textbf{52.6} & \textbf{37.8} & \textbf{23.1} & \textbf{51.5} & 27.2 & 45.1 & \textbf{59.7} & & \textbf{40.1} \\
\bottomrule
\end{tabular}}
\vspace{-3pt}
\caption{\textbf{Performance comparison of \skerl{} against baseline post-training methods} on the nine domains of \bench{}, reported as per-domain accuracy and overall \bench{} score. Across both Qwen3-4B-Instruct and Qwen3-1.7B, \skerl{} achieves the strongest overall results.
\vspace{10pt}
}
\label{tab:skerl-main}
\end{table}

\begin{figure}[t]
    \centering
    \vspace{-15pt}
    \includegraphics[width=\linewidth]{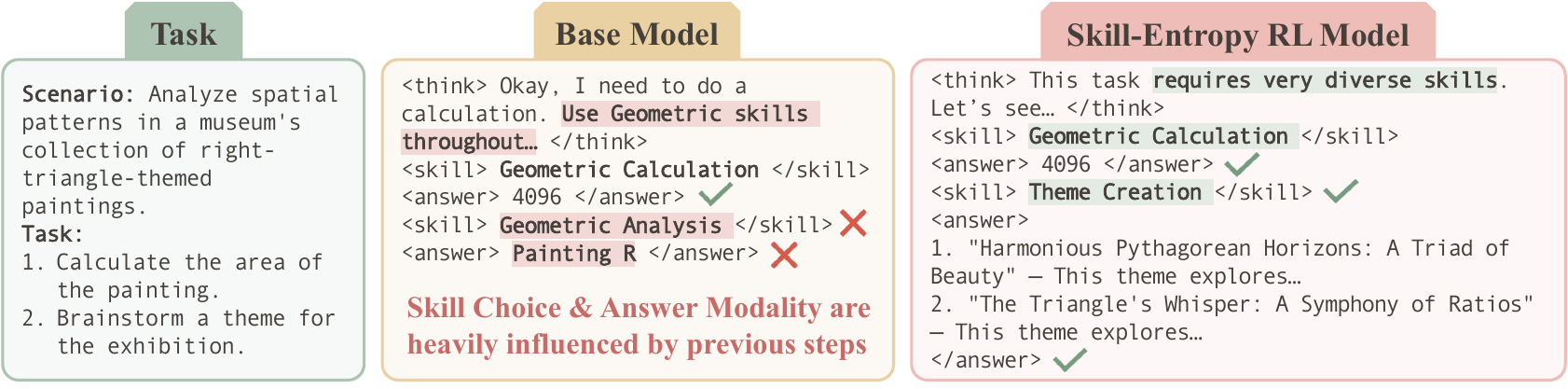}
    \caption{\textbf{Case study: \skerl{} enables cleaner skill and answer-modality switching across steps.} On this two-step task, the base model's skill choice in the second step is heavily influenced by the previous step. It reuses a math skill with a short numeric answer instead of switching to Theme Creation skill. After \skerl{} training, the model cleanly switches skills and answer modality between steps.}
    \label{fig:case-study}
    \vspace{-5pt}
\end{figure}

\subsection{Experimental Setup}
\label{sec:skerl_setup}

\textbf{Training data.}
Training uses 9K cross-skill tasks synthesized from the six verifiable
domains of \bench{} via the \bench{} pipeline
(\Cref{sec:synthesis}), with 3K used for SFT warm-up (skill-annotated
traces from a Qwen3-8B teacher) and the remaining 6K for RL; the three
open-ended domains appear only at evaluation, which follows the
Cross-Skill protocol of \Cref{sec:evaluation} on the same 300-task
test pool. Full data construction and evaluation details are in
\Cref{app:skerl_data_eval}.

\textbf{Models \& training configurations.}
We apply our method to Qwen3-4B-Instruct and Qwen3-1.7B. SFT runs for 4 epochs at learning rate 1e-5. For RL, we use GRPO with
group size $8$, prompt batch size $256$, learning rate 1e-6, KL
coefficient $10^{-3}$, and clip ratio $0.2$, on $8\!\times\!$H100s. We
weight the answer and skill-entropy rewards in \Cref{eq:skerl-reward}
as $\lambda_{\text{ans}}=0.7$ and $\lambda_{\text{ent}}=0.3$, and
ablate this choice in \Cref{app:skerl_lambda_ablation}. Full
hyperparameters and the SFT prompt template are in
\Cref{app:skerl_training,app:skerl_format}.

\textbf{Baselines.}
We compare our method against three \emph{vanilla} baselines that share
our data: (i) the \emph{base model} (no fine-tuning), (ii) \emph{SFT}
on the same 3K traces, and (iii) \emph{GRPO}, which trains on the same
6K tasks with the answer reward alone (i.e., our method without the
skill-entropy reward). All three see exactly the same prompts as our
method, so the GRPO row in \Cref{tab:skerl-main} also acts as the
ablation of our method without the skill-entropy reward. We further
compare against three \emph{skill-aware} post-training methods:
\emph{Skill-Distill}~\citep{zhang2026skilldistill}, which selects
skill-diverse traces for distillation-based fine-tuning;
\emph{SkillRL}~\citep{xia2026skillrl}, which augments RL with a
recursive skill-augmented signal; and \emph{STAT}~\citep{he2025stat},
which adapts training data toward the model's weakest skills.

\subsection{Results}

\textbf{\skerl{} achieves the best overall \bench{} performance.}
As shown in \Cref{tab:skerl-main}, \skerl{} attains the highest overall
\bench{} score on both Qwen3-4B-Instruct and Qwen3-1.7B, beating GRPO
by $+9.6\%$ and $+7.9\%$ respectively, and the strongest skill-aware
baseline (STAT) by $+7.0\%$ and $+7.1\%$. Since GRPO is exactly the
ablation of \skerl{} without the skill-entropy reward, the gap over
GRPO isolates the contribution of grading the predicted skill
structure beyond the final-answer signal alone. The cross-skill score of
\skerl{} also exceeds the base model's single-skill score
(\Cref{tab:skerl-oracle}, \Cref{app:skerl_oracle}), showing that
\skerl{} improves the underlying skill ability rather than only the
cross-skill format. A reward-weight sweep around our default
$(\lambda_{\text{ans}}, \lambda_{\text{ent}}) = (0.7, 0.3)$ is in
\Cref{app:skerl_lambda_ablation}, and results on two additional base
models, Llama-3.2-3B-Instruct and Olmo3-7B-Instruct, are reported in
\Cref{app:skerl_additional_models} and follow the same trend.

\textbf{Gains span both verifiable and open-ended domains.}
\skerl{} attains the best per-domain accuracy on seven of nine \bench{}
domains under both model sizes, with the largest gain on Creative
Writing. The open-ended gains arise even though RL training is
restricted to the six verifiable domains, suggesting that the
skill-entropy reward transfers beyond the RL distribution. The
transfer extends further to five external long-horizon and general
reasoning benchmarks outside \bench{} (MuSR, LongBench-MuSiQue,
GPQA-Diamond, MMLU, IFEval), where \skerl{} attains the highest
average score at both model sizes
(\Cref{app:skerl_external_benchmarks}).

\textbf{Case study: \skerl{} enables clean skill switches across steps.}
\Cref{fig:case-study} illustrates where the skill-entropy reward helps
on a two-step task. The base model carries its first-step math skill
over to the second step instead of switching to the Theme Creation skill
the second step requires. After \skerl{} training, the model cleanly
switches skills and answer modality between steps, matching the gold
sequence.


\subsection{Plugging Skill Entropy into Off-the-Shelf Training Data}
\label{sec:skerl_plugin}

So far we have trained on cross-skill tasks where the gold skill
sequence is available by construction. We now show that the same
pipeline plugs directly into off-the-shelf training data whose problems
were not built around an explicit skill structure (the setting
illustrated in \Cref{fig:main-figure}), making the skill-entropy reward
easy to apply to any existing training set.

\begin{wrapfigure}{r}{0.35\textwidth}
  \vspace{-15pt}
  \centering
  \captionsetup{font=footnotesize}
  \includegraphics[
    width=0.34\textwidth,
    trim=0.25cm 0.25cm 0.3cm 1cm,
    clip
  ]{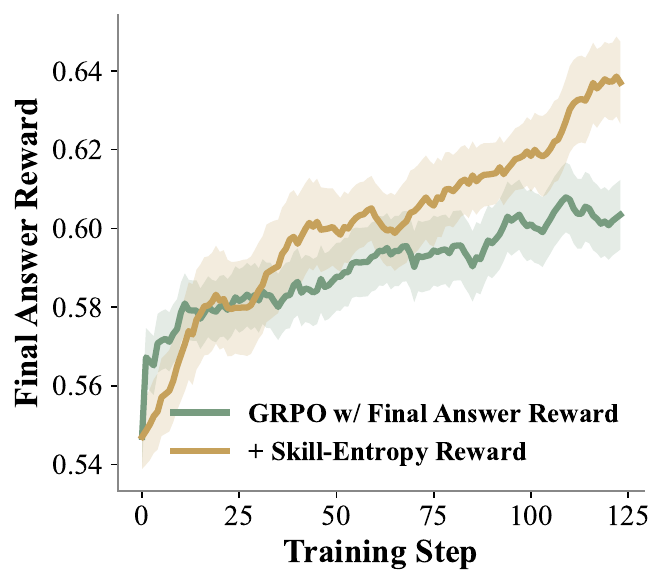}
  \vspace{-10pt}
  \caption{\textbf{Comparison of training curve on OpenR1-Math.} Adding the skill-entropy reward on top of GRPO keeps improving the reward after vanilla GRPO plateaus on \qwen{}.}
  \label{fig:evolution}
  \vspace{-20pt}
\end{wrapfigure}

\textbf{Constructing cross-skill tasks from off-the-shelf data.}
We start with OpenR1-Math~\citep{openr1}, a math-reasoning dataset with
verified answers and gold reasoning traces. We prompt an annotator
(Qwen3-8B) to segment each trace into steps and label every step with
a skill from the \bench{} skill bank (\Cref{app:skill_bank}). The
labeled sequence feeds \Cref{eq:task_entropy} to give the task-level
skill entropy; per-step intermediate conclusions serve as gold
answers, with the original final answer kept for the last step. We
train Qwen3-4B-Instruct on a 6K labeled subset under the same SFT and
RL pipeline as \Cref{sec:skerl_setup}; a worked example is in
\Cref{app:skerl_offshelf_example}.

\textbf{Results.}
\Cref{fig:evolution} plots the final-answer reward over the course of
RL. Vanilla GRPO improves slightly faster in early steps but soon
plateaus, while adding the skill-entropy reward keeps improving the
reward without saturating. The downstream numbers in
\Cref{tab:skerl-openr1} (\Cref{app:skerl_offshelf}) follow this trend:
\skerl{} attains the best score on every one of six math benchmarks,
improving over GRPO by $+1.9\%$ on average and over the base model by
$+7.7\%$. The skill-entropy reward thus delivers immediate gains on
existing training data without changing the data pipeline.





\section{Discussion}
\label{sec:discussion}
\textbf{Related Works.}
\label{sec:related_works}
We provide a more detailed discussion in \Cref{app:related_works}. Prior work falls into three directions. First, skill-aware analyses use skill annotations to compose, diagnose, or augment training data for stronger compositional reasoning~\citep{yu2023skillmix,zhao2024skillcomposition,kaur2024instructskillmix,didolkar2024metacognitive,he2025adaptmi,he2025stat,wei2026steps,jiang2025drp,zhang2026skillawaredataselectionfinetuning,chen2025symbolicmoe}, but inject the skill signal at the data or routing layer rather than into the reward of an RL stage. Second, reasoning benchmarks score a single skill in isolation~\citep{cobbe2021gsm8k,hendrycks2021math,wang2024mmlupro,suzgun2022bbh,jain2024livecodebench,lin2024zebralogic,zheng2024naturalplan,kazemi2025bbeh} or chain many turns inside an agentic environment~\citep{mialon2023gaia,liu2023agentbench,wang2025odysseybench,xu2025herobench,ye2025longproc,alazraki2025agentcoma}, but neither isolates how hard the skill switches inside a task are. Third, RL with verifiable rewards densifies the binary correctness signal through step-level or meta-reasoning supervision~\citep{shao2024deepseekmath,guo2025deepseekr1,lightman2023prm,wang2024mathshepherd,zhang2025prmsurvey,he2026sdzero,li2025multiturn,zhou2025rlvmr,yuan2025rlcompose}, but treats this difficulty as within-step rather than between-skill. \bench{} and Skill-Entropy RL fill these gaps with a single object, \emph{skill entropy}, a directed pairwise score that calibrates the benchmark and supplies an orthogonal between-step training signal.

\textbf{Conclusion.}
We introduce \emph{skill entropy}, a directed pairwise measure of how
hard it is to switch between skills inside a long-horizon task, and
use it to build \bench{}, a benchmark over $558$ skills across nine
domains that exposes a skill-switching gap invisible to single-domain
evaluations. Turning the same signal into a reward, Skill-Entropy RL
improves Qwen3-4B-Instruct's \bench{} score from $34.4\%$ to $68.4\%$,
transfers to open-ended domains, and plugs into off-the-shelf
training data such as OpenR1-Math, showing that skill entropy is
useful both as a benchmark scale and as a training signal. Our
analysis further traces this gap to a concrete failure mode: at later
steps, models tend to reuse the previous step's skill instead of
switching to the one the current step requires. 

\newpage
\bibliography{neurips_2026}

@article{hendrycks2021math,
  title={Measuring Mathematical Problem Solving with the {MATH} Dataset},
  author={Hendrycks, Dan and Burns, Collin and Kadavath, Saurav and Arora, Akul and Basart, Steven and Tang, Eric and Song, Dawn and Steinhardt, Jacob},
  journal={arXiv preprint arXiv:2103.03874},
  year={2021}
}

@article{cobbe2021gsm8k,
  title={Training Verifiers to Solve Math Word Problems},
  author={Cobbe, Karl and Kosaraju, Vineet and Bavarian, Mohammad and Chen, Mark and Jun, Heewoo and Kaiser, Lukasz and Plappert, Matthias and Tworek, Jerry and Hilton, Jacob and Nakano, Reiichiro and Hesse, Christopher and Schulman, John},
  journal={arXiv preprint arXiv:2110.14168},
  year={2021}
}

@article{wang2024mmlupro,
  title={{MMLU-Pro}: A More Robust and Challenging Multi-Task Language Understanding Benchmark},
  author={Wang, Yubo and Ma, Xueguang and Zhang, Ge and Ni, Yuansheng and Chandra, Abhranil and Guo, Shiguang and Ren, Weiming and Arulraj, Aaran and He, Xuan and Jiang, Ziyan and Li, Tianle and Ku, Max and Wang, Kai and Zhuang, Alex and Fan, Rongqi and Yue, Xiang and Chen, Wenhu},
  journal={arXiv preprint arXiv:2406.01574},
  year={2024}
}

@article{suzgun2022bbh,
  title={Challenging {BIG-Bench} Tasks and Whether Chain-of-Thought Can Solve Them},
  author={Suzgun, Mirac and Scales, Nathan and Sch{\"a}rli, Nathanael and Gehrmann, Sebastian and Tay, Yi and Chung, Hyung Won and Chowdhery, Aakanksha and Le, Quoc V and Chi, Ed H and Zhou, Denny and Wei, Jason},
  journal={arXiv preprint arXiv:2210.09261},
  year={2022}
}

@article{jain2024livecodebench,
  title={{LiveCodeBench}: Holistic and Contamination Free Evaluation of Large Language Models for Code},
  author={Jain, Naman and Han, King and Gu, Alex and Li, Wen-Ding and Yan, Fanjia and Zhang, Tianjun and Wang, Sida and Solar-Lezama, Armando and Sen, Koushik and Stoica, Ion},
  journal={arXiv preprint arXiv:2403.07974},
  year={2024}
}

@article{lin2024zebralogic,
  title={{ZebraLogic}: On the Scaling Limits of {LLMs} for Logical Reasoning},
  author={Lin, Bill Yuchen and Bras, Ronan Le and Richardson, Kyle and Sabharwal, Ashish and Poovendran, Radha and Clark, Peter and Choi, Yejin},
  journal={arXiv preprint arXiv:2502.01100},
  year={2025}
}

@article{pasupat2015wikitable,
  title={Compositional Semantic Parsing on Semi-Structured Tables},
  author={Pasupat, Panupong and Liang, Percy},
  journal={arXiv preprint arXiv:1508.00305},
  year={2015}
}

@article{chen2021websrc,
  title={{WebSRC}: A Dataset for Web-Based Structural Reading Comprehension},
  author={Chen, Xingyu and Zhao, Zihan and Chen, Lu and Zhang, Danyang and Ji, Jiabao and Luo, Ao and Xiong, Yuxuan and Yu, Kai},
  journal={arXiv preprint arXiv:2101.09465},
  year={2021}
}

@article{zheng2024naturalplan,
  title={{NATURAL PLAN}: Benchmarking {LLMs} on Natural Language Planning},
  author={Zheng, Huaixiu Steven and Mishra, Swaroop and Zhang, Hugh and Chen, Xinyun and Chen, Minmin and Nova, Azade and Hou, Le and Cheng, Heng-Tze and Le, Quoc V and Chi, Ed H and Zhou, Denny},
  journal={arXiv preprint arXiv:2406.04520},
  year={2024}
}

@article{mialon2023gaia,
  title={{GAIA}: A Benchmark for General {AI} Assistants},
  author={Mialon, Gr{\'e}goire and Fourrier, Cl{\'e}mentine and Swift, Craig and Wolf, Thomas and LeCun, Yann and Scialom, Thomas},
  journal={arXiv preprint arXiv:2311.12983},
  year={2023}
}

@article{liu2023agentbench,
  title={{AgentBench}: Evaluating {LLMs} as Agents},
  author={Liu, Xiao and Yu, Hao and Zhang, Hanchen and Xu, Yifan and Lei, Xuanyu and Lai, Hanyu and Gu, Yu and Ding, Hangliang and Men, Kaiwen and Yang, Kejuan and Zhang, Shudan and Deng, Xiang and Zeng, Aohan and Du, Zhengxiao and Zhang, Chenhui and Shen, Sheng and Zhang, Tianjun and Su, Yu and Sun, Huan and Huang, Minlie and Dong, Yuxiao and Tang, Jie},
  journal={arXiv preprint arXiv:2308.03688},
  year={2023}
}

@article{wang2025odysseybench,
  title={{OdysseyBench}: Evaluating {LLM} Agents on Long-Horizon Complex Office Application Workflows},
  author={Wang, Weixuan and Han, Dongge and Diaz, Daniel Madrigal and Xu, Jin and R{\"u}hle, Victor and Rajmohan, Saravan},
  journal={arXiv preprint arXiv:2508.09124},
  year={2025}
}

@article{xu2025herobench,
  title={{HeroBench}: A Benchmark for Long-Horizon Planning and Structured Reasoning in Virtual Worlds},
  author={Anokhin, Petr and Khalikov, Roman and Rebrikov, Stefan and Volkov, Viktor and Sorokin, Artyom and Bissonnette, Vincent},
  journal={arXiv preprint arXiv:2508.12782},
  year={2025}
}

@article{yu2023skillmix,
  title={{Skill-Mix}: A Flexible and Expandable Family of Evaluations for {AI} Models},
  author={Yu, Dingli and Kaur, Simran and Gupta, Arushi and Brown-Cohen, Jonah and Goyal, Anirudh and Arora, Sanjeev},
  journal={arXiv preprint arXiv:2310.17567},
  year={2023}
}

@article{zhao2024skillcomposition,
  title={Can Models Learn Skill Composition from Examples?},
  author={Zhao, Haoyu and Kaur, Simran and Yu, Dingli and Goyal, Anirudh and Arora, Sanjeev},
  journal={arXiv preprint arXiv:2409.19808},
  year={2024}
}

@article{shah2024aiassistedmath,
  title={{AI}-Assisted Generation of Difficult Math Questions},
  author={Shah, Vedant and Yu, Dingli and Lyu, Kaifeng and Park, Simon and Yu, Jiatong and He, Yinghui and Ke, Nan Rosemary and Mozer, Michael and Bengio, Yoshua and Arora, Sanjeev and Goyal, Anirudh},
  journal={arXiv preprint arXiv:2407.21009},
  year={2024}
}

@article{ye2025longproc,
  title={{LongProc}: Benchmarking Long-Context Language Models on Long Procedural Generation},
  author={Ye, Xi and Yin, Fangcong and He, Yinghui and Zhang, Joie and Yen, Howard and Gao, Tianyu and Durrett, Greg and Chen, Danqi},
  journal={arXiv preprint arXiv:2501.05414},
  year={2025}
}

@article{he2025adaptmi,
  title={{AdaptMI}: Adaptive Skill-based In-context Math Instruction for Small Language Models},
  author={He, Yinghui and Panigrahi, Abhishek and Lin, Yong and Arora, Sanjeev},
  journal={arXiv preprint arXiv:2505.00147},
  year={2025}
}

@article{he2025stat,
  title={{STAT}: Skill-Targeted Adaptive Training},
  author={He, Yinghui and Panigrahi, Abhishek and Lin, Yong and Arora, Sanjeev},
  journal={arXiv preprint arXiv:2510.10023},
  year={2025}
}

@article{he2026sdzero,
  title={Self-Distillation Zero: Self-Revision Turns Binary Rewards into Dense Supervision},
  author={He, Yinghui and Kaur, Simran and Bhaskar, Adithya and Yang, Yongjin and Liu, Jiarui and Ri, Narutatsu and Fowl, Liam and Panigrahi, Abhishek and Chen, Danqi and Arora, Sanjeev},
  journal={arXiv preprint arXiv:2604.12002},
  year={2026}
}

@article{guo2025deepseekr1,
  title={{DeepSeek-R1}: Incentivizing Reasoning Capability in {LLMs} via Reinforcement Learning},
  author={Guo, Daya and Yang, Dejian and Zhang, Haowei and Song, Junxiao and Zhang, Ruoyu and Xu, Runxin and Zhu, Qihao and Ma, Shirong and Wang, Peiyi and Bi, Xiao and others},
  journal={arXiv preprint arXiv:2501.12948},
  year={2025}
}

@article{shao2024deepseekmath,
  title={{DeepSeekMath}: Pushing the Limits of Mathematical Reasoning in Open Language Models},
  author={Shao, Zhihong and Wang, Peiyi and Zhu, Qihao and Xu, Runxin and Song, Junxiao and Bi, Xiao and Zhang, Haowei and Zhang, Mingchuan and Li, Y. K. and Wu, Y. and Guo, Daya},
  journal={arXiv preprint arXiv:2402.03300},
  year={2024}
}

@article{lightman2023prm,
  title={Let's Verify Step by Step},
  author={Lightman, Hunter and Kosaraju, Vineet and Burda, Yura and Edwards, Harri and Baker, Bowen and Lee, Teddy and Leike, Jan and Schulman, John and Sutskever, Ilya and Cobbe, Karl},
  journal={arXiv preprint arXiv:2305.20050},
  year={2023}
}

@article{wang2024mathshepherd,
  title={{Math-Shepherd}: Verify and Reinforce {LLMs} Step-by-step without Human Annotations},
  author={Wang, Peiyi and Li, Lei and Shao, Zhihong and Xu, R. X. and Dai, Damai and Li, Yifei and Chen, Deli and Wu, Y. and Sui, Zhifang},
  journal={arXiv preprint arXiv:2312.08935},
  year={2024}
}

@article{zhang2025prmsurvey,
  title={A Survey of Process Reward Models: From Outcome Signals to Process Supervisions for Large Language Models},
  author={Zheng, Congmin and Zhu, Jiachen and Ou, Zhuoying and Chen, Yuxiang and Zhang, Kangning and Shan, Rong and Zheng, Zeyu and Yang, Mengyue and Lin, Jianghao and Yu, Yong and Zhang, Weinan},
  journal={arXiv preprint arXiv:2510.08049},
  year={2025}
}

@article{openr1,
  title={{Open-R1}: A Fully Open Reproduction of {DeepSeek-R1}},
  author={{Hugging Face}},
  journal={https://github.com/huggingface/open-r1},
  year={2025}
}

@article{cheng2025guru,
  title={Revisiting Reinforcement Learning for {LLM} Reasoning from A Cross-Domain Perspective},
  author={Cheng, Zhoujun and Hao, Shibo and Liu, Tianyang and Zhou, Fan and Xie, Yutao and Yao, Feng and Bian, Yuexin and Zhuang, Yonghao and Dey, Nilabjo and Zha, Yuheng and Gu, Yi and Zhou, Kun and Wang, Yuqi and Li, Yuan and Fan, Richard and She, Jianshu and Gao, Chengqian and Saparov, Abulhair and Li, Haonan and Killian, Taylor W. and Yurochkin, Mikhail and Liu, Zhengzhong and Xing, Eric P. and Hu, Zhiting},
  journal={arXiv preprint arXiv:2506.14965},
  year={2025}
}

@misc{zhang2026skillawaredataselectionfinetuning,
      title={Skill-Aware Data Selection and Fine-Tuning for Data-Efficient Reasoning Distillation},
      author={Zhang, Lechen and Zhang, Yunxiang and Hu, Wei and Wang, Lu},
      year={2026},
      eprint={2601.10109},
      archivePrefix={arXiv},
      primaryClass={cs.CL},
      url={https://arxiv.org/abs/2601.10109},
}

@article{kazemi2025bbeh,
  title={{BIG-Bench Extra Hard}},
  author={Kazemi, Mehran and Fatemi, Bahare and Bansal, Hritik and Palowitch, John and Anastasiou, Chrysovalantis and Mehta, Sanket Vaibhav and Jain, Lalit K. and Aglietti, Virginia and Jindal, Disha and Chen, Peter and Dikkala, Nishanth and Tyen, Gladys and Liu, Xin and Shalit, Uri and Chiappa, Silvia and Olszewska, Kate and Tay, Yi and Tran, Vinh Q. and Le, Quoc V. and Firat, Orhan},
  journal={arXiv preprint arXiv:2502.19187},
  year={2025}
}

@article{kaur2024instructskillmix,
  title={{Instruct-SkillMix}: A Powerful Pipeline for {LLM} Instruction Tuning},
  author={Kaur, Simran and Park, Simon and Goyal, Anirudh and Arora, Sanjeev},
  journal={arXiv preprint arXiv:2408.14774},
  year={2024}
}

@article{zhou2025rlvmr,
  title={{RLVMR}: Reinforcement Learning with Verifiable Meta-Reasoning Rewards for Robust Long-Horizon Agents},
  author={Zhang, Zijing and Chen, Ziyang and Li, Mingxiao and Tu, Zhaopeng and Li, Xiaolong},
  journal={arXiv preprint arXiv:2507.22844},
  year={2025}
}

@article{li2025multiturn,
  title={Reinforcing Multi-Turn Reasoning in {LLM} Agents via Turn-Level Reward Design},
  author={Wei, Quan and Zeng, Siliang and Li, Chenliang and Brown, William and Frunza, Oana and Deng, Wei and Schneider, Anderson and Nevmyvaka, Yuriy and Zhao, Yang Katie and Garcia, Alfredo and Hong, Mingyi},
  journal={arXiv preprint arXiv:2505.11821},
  year={2025}
}

@article{didolkar2024metacognitive,
  title={Metacognitive Capabilities of {LLMs}: An Exploration in Mathematical Problem Solving},
  author={Didolkar, Aniket and Goyal, Anirudh and Ke, Nan Rosemary and Guo, Siyuan and Valko, Michal and Lillicrap, Timothy and Rezende, Danilo and Bengio, Yoshua and Mozer, Michael and Arora, Sanjeev},
  journal={Advances in Neural Information Processing Systems},
  year={2024}
}

@article{yuan2025rlcompose,
  title={From $f(x)$ and $g(x)$ to $f(g(x))$: {LLMs} Learn New Skills in {RL} by Composing Old Ones},
  author={Yuan, Lifan and Chen, Weize and Zhang, Yuchen and Cui, Ganqu and Wang, Hanbin and You, Ziming and Ding, Ning and Liu, Zhiyuan and Sun, Maosong and Peng, Hao},
  journal={arXiv preprint arXiv:2509.25123},
  year={2025}
}

@article{alazraki2025agentcoma,
  title={{AgentCoMa}: A Compositional Benchmark Mixing Commonsense and Mathematical Reasoning in Real-World Scenarios},
  author={Alazraki, Lisa and Chen, Lihu and Brassard, Ana and Stacey, Joe and Rahmani, Hossein A. and Rei, Marek},
  journal={arXiv preprint arXiv:2508.19988},
  year={2025}
}

@article{chen2025symbolicmoe,
  title={Symbolic Mixture-of-Experts: Adaptive Skill-based Routing for Heterogeneous Reasoning},
  author={Chen, Justin Chih-Yao and Yun, Sukwon and Stengel-Eskin, Elias and Chen, Tianlong and Bansal, Mohit},
  journal={arXiv preprint arXiv:2503.05641},
  year={2025}
}

@article{wei2026steps,
  title={Towards Compositional Generalization of {LLMs} via Skill Taxonomy Guided Data Synthesis},
  author={Wei, Yifan and Du, Li and Yu, Xiaoyan and Feng, Yang and Li, Angsheng},
  journal={arXiv preprint arXiv:2601.03676},
  year={2026}
}

@article{jiang2025drp,
  title={{DRP}: Distilled Reasoning Pruning with Skill-aware Step Decomposition for Efficient Large Reasoning Models},
  author={Jiang, Yuxuan and Li, Dawei and Ferraro, Francis},
  journal={arXiv preprint arXiv:2505.13975},
  year={2025}
}

@article{li2025skillsbench,
  title={{SkillsBench}: Benchmarking How Well Agent Skills Work Across Diverse Tasks},
  author={Li, Xiangyi and Chen, Wenbo and Liu, Yimin and Zheng, Shenghan and Chen, Xiaokun and He, Yifeng and Li, Yubo and You, Bingran and Shen, Haotian and Sun, Jiankai and Wang, Shuyi and Li, Binxu and Zeng, Qunhong and Wang, Di and Zhao, Xuandong and Wang, Yuanli and Ben Chaim, Roey and Di, Zonglin and Gao, Yipeng and He, Junwei and He, Yizhuo and Jing, Liqiang and Kong, Luyang and Lan, Xin and Li, Jiachen and Li, Songlin and Li, Yijiang and Lin, Yueqian and Liu, Xinyi and Liu, Xuanqing and Lyu, Haoran and Ma, Ze and Wang, Bowei and Wang, Runhui and Wang, Tianyu and Ye, Wengao and Zhang, Yue and Xing, Hanwen and Xue, Yiqi and Dillmann, Steven and Lee, Han-chung},
  journal={arXiv preprint arXiv:2602.12670},
  year={2026}
}

@article{zhong2025skilllearnbench,
  title={{SkillLearnBench}: Benchmarking Continual Learning Methods for Agent Skill Generation on Real-World Tasks},
  author={Zhong, Shanshan and Lu, Yi and Ning, Jingjie and Wan, Yibing and Feng, Lihan and Ao, Yuyi and Ribeiro, Leonardo F. R. and Dreyer, Markus and Ammirati, Sean and Xiong, Chenyan},
  journal={arXiv preprint arXiv:2604.20087},
  year={2026}
}

@article{retrieverank2025skillselection,
  title={{SkillRouter}: Skill Routing for {LLM} Agents at Scale},
  author={Zheng, YanZhao and Zhang, ZhenTao and Ma, Chao and Yu, YuanQiang and Zhu, JiHuai and Wu, Yong and Xu, Tianze and Dong, Baohua and Zhu, Hangcheng and Huang, Ruohui and Yu, Gang},
  journal={arXiv preprint arXiv:2603.22455},
  year={2026}
}

@article{chen2025skillcraft,
  title={{SkillCraft}: Can {LLM} Agents Learn to Use Tools Skillfully?},
  author={Chen, Shiqi and Gai, Jingze and Zhou, Ruochen and Zhang, Jinghan and Zhu, Tongyao and Li, Junlong and Wang, Kangrui and Wang, Zihan and Chen, Zhengyu and Kaleb, Klara and Miao, Ning and Gao, Siyang and Lu, Cong and Li, Manling and He, Junxian and Teh, Yee Whye},
  journal={arXiv preprint arXiv:2603.00718},
  year={2026}
}

@article{an2025stad,
  title={{STaD}: Scaffolded Task Design for Identifying Compositional Skill Gaps in {LLMs}},
  author={An, Sungeun and Kadhe, Swanand Ravindra and Thakur, Shailja and DeLuca, Chad and Patel, Hima},
  journal={arXiv preprint arXiv:2604.18177},
  year={2026}
}

@article{logicskills2025,
  title={{LogicSkills}: A Structured Benchmark for Formal Reasoning in Large Language Models},
  author={Rabern, Brian and Mondorf, Philipp and Plank, Barbara},
  journal={arXiv preprint arXiv:2602.06533},
  year={2026}
}

@article{latentskill2024cot,
  title={{LaRS}: Latent Reasoning Skills for Chain-of-Thought Reasoning},
  author={Xu, Zifan and Wang, Haozhu and Bespalov, Dmitriy and Wu, Xian and Stone, Peter and Qi, Yanjun},
  journal={arXiv preprint arXiv:2312.04684},
  year={2024}
}

@article{concepts2025multimodal,
  title={Concepts or Skills? Rethinking Instruction Selection for Multi-modal Models},
  author={Bai, Andrew and Cui, Justin and Wang, Ruochen and Hsieh, Cho-Jui},
  journal={arXiv preprint arXiv:2508.10339},
  year={2025}
}

@article{wang2025skilllibrary,
  title={Reinforcement Learning for Self-Improving Agent with Skill Library},
  author={Wang, Jiongxiao and Yan, Qiaojing and Wang, Yawei and Tian, Yijun and Mishra, Soumya Smruti and Xu, Zhichao and Gandhi, Megha and Xu, Panpan and Cheong, Lin Lee},
  journal={arXiv preprint arXiv:2512.17102},
  year={2025}
}

@article{hu2025memoryagentbench,
  title={Evaluating Memory in {LLM} Agents via Incremental Multi-Turn Interactions},
  author={Hu, Yuanzhe and Wang, Yu and McAuley, Julian},
  journal={arXiv preprint arXiv:2507.05257},
  year={2025}
}

@article{sakai2025orderedcommongen,
  title={Revisiting Compositional Generalization Capability of Large Language Models Considering Instruction Following Ability},
  author={Sakai, Yusuke and Kamigaito, Hidetaka and Watanabe, Taro},
  journal={arXiv preprint arXiv:2506.15629},
  year={2025}
}

@article{compositionalrelational2025,
  title={Benchmarking and Understanding Compositional Relational Reasoning of {LLMs}},
  author={Ni, Ruikang and Xiao, Da and Meng, Qingye and Li, Xiangyu and Zheng, Shihui and Liang, Hongliang},
  journal={arXiv preprint arXiv:2412.12841},
  year={2024}
}

@article{generalagentbench2025,
  title={Benchmark Test-Time Scaling of General {LLM} Agents},
  author={Li, Xiaochuan and Ming, Ryan and Setlur, Pranav and Paladugu, Abhijay and Tang, Andy and Kang, Hao and Shao, Shuai and Jin, Rong and Xiong, Chenyan},
  journal={arXiv preprint arXiv:2602.18998},
  year={2026}
}

@article{mtrbench2025,
  title={{MTR-Bench}: A Comprehensive Benchmark for Multi-Turn Reasoning Evaluation},
  author={Li, Xiaoyuan and Bao, Keqin and Ma, Yubo and Li, Moxin and Wang, Wenjie and Men, Rui and Zhang, Yichang and Feng, Fuli and Liu, Dayiheng and Lin, Junyang},
  journal={arXiv preprint arXiv:2505.17123},
  year={2025}
}

@article{mscore2025,
  title={{MSCoRe}: A Benchmark for Multi-Stage Collaborative Reasoning in {LLM} Agents},
  author={Lei, Yuzhen and Xie, Hongbin and Zhao, Jiaxing and Liu, Shuangxue and Song, Xuan},
  journal={arXiv preprint arXiv:2509.17628},
  year={2025}
}

@article{benchprofiling2025,
  title={Benchmark Profiling: Mechanistic Diagnosis of {LLM} Benchmarks},
  author={Kim, Dongjun and Shim, Gyuho and Chun, Yongchan and Kim, Minhyuk and Park, Chanjun and Lim, Heuiseok},
  journal={arXiv preprint arXiv:2510.01232},
  year={2025}
}

@article{guo2025rbench,
  title={{R-Bench}: Graduate-level Multi-disciplinary Benchmarks for {LLM} \& {MLLM} Complex Reasoning Evaluation},
  author={Guo, Meng-Hao and Xu, Jiajun and Zhang, Yi and Song, Jiaxi and Peng, Haoyang and Deng, Yi-Xuan and Dong, Xinzhi and Nakayama, Kiyohiro and Geng, Zhengyang and Wang, Chen and Ni, Bolin and Yang, Guo-Wei and Rao, Yongming and Peng, Houwen and Hu, Han and Wetzstein, Gordon and Hu, Shi-min},
  journal={arXiv preprint arXiv:2505.02018},
  year={2025}
}

@article{shu2025darebench,
  title={{DARE-bench}: Evaluating Modeling and Instruction Fidelity of {LLMs} in Data Science},
  author={Shu, Fan and Wang, Yite and Wu, Ruofan and Liu, Boyi and Yao, Zhewei and He, Yuxiong and Yan, Feng},
  journal={arXiv preprint arXiv:2602.24288},
  year={2026}
}

@article{paperbench2025,
  title={{PaperBench}: Evaluating {AI}'s Ability to Replicate {AI} Research},
  author={Starace, Giulio and Jaffe, Oliver and Sherburn, Dane and Aung, James and Chan, Jun Shern and Maksin, Leon and Dias, Rachel and Mays, Evan and Kinsella, Benjamin and Thompson, Wyatt and Heidecke, Johannes and Glaese, Amelia and Patwardhan, Tejal},
  journal={arXiv preprint arXiv:2504.01848},
  year={2025}
}

@article{testexploiters2025,
  title={Reasoning Models are Test Exploiters: Rethinking Multiple-Choice},
  author={Raman, Narun and Lundy, Taylor and Leyton-Brown, Kevin},
  journal={arXiv preprint arXiv:2507.15337},
  year={2025}
}

@article{chen2025selfevolvingcurriculum,
  title={Self-Evolving Curriculum for {LLM} Reasoning},
  author={Chen, Xiaoyin and Lu, Jiarui and Kim, Minsu and Zhang, Dinghuai and Tang, Jian and Piche, Alexandre and Gontier, Nicolas and Bengio, Yoshua and Kamalloo, Ehsan},
  journal={arXiv preprint arXiv:2505.14970},
  year={2025}
}

@article{pang2025bootstrappingmath,
  title={Reasoning Curriculum: Bootstrapping Broad {LLM} Reasoning from Math},
  author={Pang, Bo and Kong, Deqian and Savarese, Silvio and Xiong, Caiming and Zhou, Yingbo},
  journal={arXiv preprint arXiv:2510.26143},
  year={2025}
}

@article{fang2025thinkless,
  title={Thinkless: {LLM} Learns When to Think},
  author={Fang, Gongfan and Ma, Xinyin and Wang, Xinchao},
  journal={arXiv preprint arXiv:2505.13379},
  year={2025}
}

@article{li2025start,
  title={{START}: Self-taught Reasoner with Tools},
  author={Li, Chengpeng and Xue, Mingfeng and Zhang, Zhenru and Yang, Jiaxi and Zhang, Beichen and Wang, Xiang and Yu, Bowen and Hui, Binyuan and Lin, Junyang and Liu, Dayiheng},
  journal={arXiv preprint arXiv:2503.04625},
  year={2025}
}

@article{yang2025deepcritic,
  title={{DeepCritic}: Deliberate Critique with Large Language Models},
  author={Yang, Wenkai and Chen, Jingwen and Lin, Yankai and Wen, Ji-Rong},
  journal={arXiv preprint arXiv:2505.00662},
  year={2025}
}

@article{zhang2025graphofthought,
  title={{DAG-Math}: Graph-of-Thought Guided Mathematical Reasoning in {LLMs}},
  author={Zhang, Yuanhe and Kuzborskij, Ilja and Lee, Jason D. and Leng, Chenlei and Liu, Fanghui},
  journal={arXiv preprint arXiv:2510.19842},
  year={2025}
}

@article{qi2025parallelsamples,
  title={Learning to Reason Across Parallel Samples for {LLM} Reasoning},
  author={Qi, Jianing and Ye, Xi and Tang, Hao and Zhu, Zhigang and Choi, Eunsol},
  journal={arXiv preprint arXiv:2506.09014},
  year={2025}
}

@inproceedings{zhang2026skilldistill,
  title={Skill-Aware Data Selection and Fine-Tuning for Data-Efficient Reasoning Distillation},
  author={Zhang and others},
  booktitle={Annual Meeting of the Association for Computational Linguistics (ACL)},
  year={2026}
}

@misc{xia2026skillrl,
  title={{SkillRL}: Evolving Agents via Recursive Skill-Augmented Reinforcement Learning},
  author={Xia and others},
  year={2026}
}

@inproceedings{qiu2025emoagent,
  title={{EmoAgent}: Assessing and Safeguarding Human-{AI} Interaction for Mental Health Safety},
  author={Qiu, Jiahao and He, Yinghui and Juan, Xinzhe and Wang, Yimin and Liu, Yuhan and Yao, Zixin and Wu, Yue and Jiang, Xun and Yang, Ling and Wang, Mengdi},
  booktitle={Proceedings of the 2025 Conference on Empirical Methods in Natural Language Processing (EMNLP)},
  year={2025}
}

@article{yang2026tac,
  title={Transferability for General Reasoning: An Automated Curriculum for Multi-Domain {RLVR}},
  author={Yang, Yongjin and Liu, Jiarui and He, Yinghui and Zhang, Lechen and Sch{\"o}lkopf, Bernhard and Jin, Zhijing},
  journal={arXiv preprint arXiv:2606.25178},
  year={2026}
}

@inproceedings{wu2023hitom,
  title={{Hi-ToM}: A Benchmark for Evaluating Higher-Order Theory of Mind Reasoning in Large Language Models},
  author={Wu, Yufan and He, Yinghui and Jia, Yilin and Mihalcea, Rada and Chen, Yulong and Deng, Naihao},
  booktitle={Findings of the Association for Computational Linguistics: EMNLP 2023},
  year={2023}
}

@inproceedings{wu2026criticl,
  title={{CritICL}: Inference-Time Weak-to-Strong Generalization from Small Language Model Failure Modes},
  author={Wu, Yufan and Hu, Zhengyi and Wei, Lang and Li, Ruichen and Yang, Qifan and He, Yinghui and Zhu, Ting},
  booktitle={Conference on Language Modeling (COLM)},
  year={2026}
}
\bibliographystyle{plainnat}

\newpage
\appendix
\section*{Appendix Content}
\vspace{0.5em}
\noindent
\hyperref[app:related_works]{A \hspace{0.5em} Related Works} \dotfill \pageref{app:related_works} \\[6pt]
\hyperref[app:synthesis]{B \hspace{0.5em} Synthesis Pipeline Details} \dotfill \pageref{app:synthesis} \\[2pt]
\hyperref[app:datasets]{\hspace{1.7em} B.1 \hspace{0.5em} Seed Datasets and Rubric Prompts} \dotfill \pageref{app:datasets} \\[2pt]
\hyperref[app:skill_bank]{\hspace{1.7em} B.2 \hspace{0.5em} Skill Bank} \dotfill \pageref{app:skill_bank} \\[2pt]
\hyperref[app:entropy_estimation]{\hspace{1.7em} B.3 \hspace{0.5em} Skill Entropy Estimation Procedure} \dotfill \pageref{app:entropy_estimation} \\[2pt]
\hyperref[app:ske_factorization]{\hspace{1.7em} B.4 \hspace{0.5em} Factorizing Pairwise Skill Entropy Through Domains} \dotfill \pageref{app:ske_factorization} \\[2pt]
\hyperref[app:reference_model]{\hspace{1.7em} B.5 \hspace{0.5em} Choice of Reference Model} \dotfill \pageref{app:reference_model} \\[2pt]
\hyperref[app:sequence_sampling]{\hspace{1.7em} B.6 \hspace{0.5em} Skill-Sequence Sampling} \dotfill \pageref{app:sequence_sampling} \\[2pt]
\hyperref[app:proposer]{\hspace{1.7em} B.7 \hspace{0.5em} Proposer and Open-Ended Generation Prompts} \dotfill \pageref{app:proposer} \\[2pt]
\hyperref[app:example]{\hspace{1.7em} B.8 \hspace{0.5em} Worked Example} \dotfill \pageref{app:example} \\[2pt]
\hyperref[app:filtering]{\hspace{1.7em} B.9 \hspace{0.5em} Filtering and Verification} \dotfill \pageref{app:filtering} \\[6pt]
\hyperref[app:full_results]{C \hspace{0.5em} Full Evaluation Results} \dotfill \pageref{app:full_results} \\[2pt]
\hyperref[app:eval_settings]{\hspace{1.7em} C.1 \hspace{0.5em} Inference Settings} \dotfill \pageref{app:eval_settings} \\[6pt]
\hyperref[app:skerl]{D \hspace{0.5em} Skill-Entropy RL Training Details} \dotfill \pageref{app:skerl} \\[2pt]
\hyperref[app:skerl_offshelf]{\hspace{1.7em} D.1 \hspace{0.5em} Off-the-Shelf Training Results on OpenR1-Math} \dotfill \pageref{app:skerl_offshelf} \\[2pt]
\hyperref[app:skerl_format]{\hspace{1.7em} D.2 \hspace{0.5em} Response Format and Example SFT Trace} \dotfill \pageref{app:skerl_format} \\[2pt]
\hyperref[app:skerl_reward]{\hspace{1.7em} D.3 \hspace{0.5em} Reward Definitions} \dotfill \pageref{app:skerl_reward} \\[2pt]
\hyperref[app:skerl_skill_match]{\hspace{1.7em} D.4 \hspace{0.5em} Embedding-Based Skill Matching} \dotfill \pageref{app:skerl_skill_match} \\[2pt]
\hyperref[app:skerl_data_eval]{\hspace{1.7em} D.5 \hspace{0.5em} Training Data and Evaluation Protocol} \dotfill \pageref{app:skerl_data_eval} \\[2pt]
\hyperref[app:skerl_training]{\hspace{1.7em} D.6 \hspace{0.5em} RL Hyperparameters} \dotfill \pageref{app:skerl_training} \\[2pt]
\hyperref[app:skerl_lambda_ablation]{\hspace{1.7em} D.7 \hspace{0.5em} Reward Weight Ablation} \dotfill \pageref{app:skerl_lambda_ablation} \\[2pt]
\hyperref[app:skerl_oracle]{\hspace{1.7em} D.8 \hspace{0.5em} Comparison Against the Single-Skill Oracle} \dotfill \pageref{app:skerl_oracle} \\[2pt]
\hyperref[app:skerl_additional_models]{\hspace{1.7em} D.9 \hspace{0.5em} Additional Base Models} \dotfill \pageref{app:skerl_additional_models} \\[2pt]
\hyperref[app:skerl_external_benchmarks]{\hspace{1.7em} D.10 \hspace{0.5em} Generalization to External Benchmarks} \dotfill \pageref{app:skerl_external_benchmarks} \\[2pt]
\hyperref[app:skerl_offshelf_example]{\hspace{1.7em} D.11 \hspace{0.5em} Worked Example: Labeling an Off-the-Shelf Problem} \dotfill \pageref{app:skerl_offshelf_example} \\

\newpage

\section{Related Works}
\label{app:related_works}

This appendix expands the discussion summarized in the main paper's Related Works (\Cref{sec:related_works}) along three directions: reasoning benchmarks, skill-aware analyses and training of LLMs, and reinforcement learning with verifiable and dense rewards.

\paragraph{Reasoning benchmarks for LLMs.}
One line of benchmarks targets a single reasoning skill in isolation and reports a single accuracy per domain, covering math, multi-domain knowledge, coding, logic, planning, graduate-level reasoning, and theory of mind~\citep{cobbe2021gsm8k,hendrycks2021math,wang2024mmlupro,suzgun2022bbh,kazemi2025bbeh,jain2024livecodebench,lin2024zebralogic,zheng2024naturalplan,guo2025rbench,wu2023hitom}, with more recent variants probing narrower competences such as data-analysis fidelity, AI-research replication, and robustness of multiple-choice evaluation~\citep{shu2025darebench,paperbench2025,testexploiters2025,benchprofiling2025}. A second line moves to long-horizon agentic settings where models must chain many turns, tool calls, or procedural steps in a complex environment~\citep{mialon2023gaia,liu2023agentbench,wang2025odysseybench,xu2025herobench,ye2025longproc,generalagentbench2025,mtrbench2025,mscore2025,hu2025memoryagentbench,qiu2025emoagent}. Closer in spirit to our setup, a third line probes skill composition inside a single passage or synthetic problem and finds sharp accuracy drops once two skills are combined~\citep{yu2023skillmix,zhao2024skillcomposition,shah2024aiassistedmath,alazraki2025agentcoma,sakai2025orderedcommongen,compositionalrelational2025,an2025stad}, while a parallel line builds skill-tagged benchmarks that score coverage over a fixed skill bank~\citep{li2025skillsbench,zhong2025skilllearnbench,logicskills2025}. None of these settings isolate how hard a particular skill switch is. \bench{} inherits the skill-centric view but builds long-horizon tasks whose successive steps each invoke a different skill, and introduces \emph{skill entropy} as a directed pairwise quantity that scores how hard a switch is for a reference model rather than treating all switches as uniform.

\paragraph{Skill-aware analysis and training of LLMs.}
A growing body of work uses skills as a structural lens on training data. Early evaluations showed that skill-tagged data can both diagnose and incentivize compositional behavior~\citep{yu2023skillmix,zhao2024skillcomposition,didolkar2024metacognitive}. Building on this view, a first thread synthesizes or selects training data along a skill axis, by sampling skill combinations, inducing skill taxonomies, retrieving skill-targeted examples, reweighting toward missing skills, pruning skill-redundant steps, or selecting compact skill-balanced subsets~\citep{kaur2024instructskillmix,wei2026steps,he2025adaptmi,he2025stat,jiang2025drp,zhang2026skillawaredataselectionfinetuning,concepts2025multimodal}. A second thread discovers or grows skills as latent objects, either induced from chain-of-thought traces or maintained as a skill library that an RL agent expands over time~\citep{latentskill2024cot,wang2025skilllibrary}. A third thread uses skills at inference time for retrieval, tool-use evaluation, and expert composition~\citep{retrieverank2025skillselection,chen2025skillcraft,chen2025symbolicmoe}. All of these inject the skill signal at the data, latent, or routing layer. We instead integrate skill structure into the reward of an RL stage: the model is graded on whether the skill chain it commits to matches the difficulty profile of the task, which transfers beyond the training distribution to open-ended domains.

\paragraph{Reinforcement learning with verifiable and dense rewards.}
RL with verifiable rewards is the standard pipeline for training reasoning models, optimizing binary correctness with GRPO and related algorithms~\citep{shao2024deepseekmath,guo2025deepseekr1,cheng2025guru}. To address reward sparsity, one line of work supplies step-level supervision through process reward models~\citep{lightman2023prm,wang2024mathshepherd,zhang2025prmsurvey}, and a closely related line densifies the outcome signal within a single solution via self-revision, critique, or parallel-sample aggregation~\citep{he2026sdzero,yang2025deepcritic,qi2025parallelsamples,wu2026criticl}. A second line designs turn-level or meta-reasoning rewards that grade intermediate cognitive steps in long-horizon agents~\citep{li2025multiturn,zhou2025rlvmr}. A third line studies what RL teaches: skill composition under suitable training tasks, transfer from math-only RL to broader reasoning, and learning when to invoke explicit reasoning~\citep{yuan2025rlcompose,pang2025bootstrappingmath,fang2025thinkless}. Orthogonal lines shape the training distribution or trajectory through self-evolving curricula, tool use, or graph-structured plans~\citep{chen2025selfevolvingcurriculum,li2025start,zhang2025graphofthought,yang2026tac}. All of these target the difficulty \emph{within} a step. Our method adds an orthogonal dense signal that targets the difficulty \emph{between} steps: a task-structural reward that grades the predicted skill sequence against the gold one through skill entropy. The two ideas compose: process- and self-revision-based rewards densify within a step, whereas skill entropy densifies across steps.

\section{Synthesis pipeline details}
\label{app:synthesis}

This appendix gives the per-stage details of the \bench{} synthesis
pipeline summarized in \Cref{sec:synthesis}: seed datasets and judges
(\Cref{app:datasets}), the skill bank (\Cref{app:skill_bank}), the
skill entropy estimation procedure (\Cref{app:entropy_estimation}),
the skill$\to$domain factorization (\Cref{app:ske_factorization}), the
choice of reference model (\Cref{app:reference_model}), skill-sequence
sampling (\Cref{app:sequence_sampling}), the proposer and open-ended
generation prompts (\Cref{app:proposer}), a worked end-to-end example
(\Cref{app:example}), and the verifier checklist plus rejection rates
(\Cref{app:filtering}).

\subsection{Seed datasets and rubric prompts}
\label{app:datasets}

This subsection details the seed datasets and per-domain scorers used in
the synthesis pipeline of \Cref{sec:synthesis} and the per-domain
evaluator $\operatorname{eval}_d$ defined in \Cref{sec:cross_skill_tasks}.
\Cref{tab:source-datasets} lists the seed dataset, license, and split
used for each domain, together with whether grading is deterministic
(verifiable domains) or rubric-based (open-ended domains). Rubric prompts
for the three open-ended domains are given inline below.

\begin{table}[h]
\centering
\small
\setlength{\tabcolsep}{4pt}
\begin{tabular}{l l l l}
\toprule
Domain & Source & Type & Scoring \\
\midrule
math                   & OpenR1-Math                            & verifiable  & numeric / symbolic equivalence \\
science                & MMLU-Pro                               & verifiable  & multiple-choice match \\
coding                 & LiveCodeBench                          & verifiable  & sandboxed unit-test execution \\
logic                  & ZebraLogicBench, guru-RL-92k           & verifiable  & grid equality \\
information extraction & WikiTableQuestions, WebSRC v1.0        & verifiable  & exact / numeric match \\
planning               & NaturalPlan                            & verifiable  & constraint satisfaction \\
Creative Writing       & LLM-generated                          & open-ended  & rubric-based LLM judge \\
Context Retrieval      & LLM-generated                          & open-ended  & rubric-based LLM judge \\
instruction following  & LLM-generated                          & open-ended  & rubric-based LLM judge \\
\bottomrule
\end{tabular}
\vspace{10pt}
\caption{Seed datasets and scoring per domain.}
\label{tab:source-datasets}
\end{table}

\paragraph{Rubric generation prompt.} For each open-ended step, the
proposer LLM is asked to emit both an open-ended question and a per-skill
rubric in a single call, conditioned on the unifying scenario, the task
plan, and the previous-step questions of the task. The rubric is then
used by the judge as a per-skill checklist together with a standard
scoring instruction (state the question, the response, and the rubric;
ask the judge to return a $[0,1]$ score and a short justification);
per-skill checklists are listed in \Cref{app:skill_bank}. The verbatim
rubric generation prompt template is shown below, and one example
question--rubric pair per open-ended domain follows.

\begin{promptlisting}{Rubric generation prompt (open-ended domains)}
You are an expert in designing open-ended questions and rubrics for long-horizon tasks. You'll be given a scenario, a task plan, and previous steps of the task, and you'll need to propose a question and a rubric for the step, based on the given domain and skill.


### Scenario: {scenario}

### Task Plan: {description}

### Previous Steps:
{previous_context}

### Current Step:
Domain: {domain}
Skill: {skill}

Please propose an open-ended Question for this step that fits the scenario and uses the Skill {skill} in the Domain {domain}, then propose a rubric for grading the answer. Note: your question should emphasize the use of Domain {domain}, requiring the Skill {skill} to solve the problem.

Output Format:
<question> [Your question] </question>
<rubric> [Your rubric] </rubric>
\end{promptlisting}

\begin{examplelisting}{Example output -- Creative Writing (skill: storytelling)}
<question>
Create a narrative that captures the potential international crisis by weaving together the key factors from the previous steps. Your story should highlight the competing nations or factions, their motivations, and the critical thresholds that could trigger the crisis. Use specific examples from the data provided to support your narrative.
</question>

<rubric>
1. Clarity and Coherence (20 points): The narrative is well-structured, easy to follow, and logically organized. It clearly outlines the key factors and their interrelationships.
2. Depth of Analysis (30 points): The narrative delves into the motivations and dynamics of the competing nations or factions, providing a nuanced understanding of the situation. Specific examples from the data are used to support the narrative.
3. Creativity and Engagement (20 points): The narrative is engaging and creative, using vivid language and storytelling techniques to bring the scenario to life. It captures the reader's interest and effectively conveys the potential crisis.
4. Relevance and Accuracy (30 points): The narrative accurately reflects the information provided in the previous steps and is relevant to the task of predicting the outcome of the international crisis. It uses the data and insights from the previous steps to construct a plausible and coherent narrative.
</rubric>
\end{examplelisting}

\begin{examplelisting}{Example output -- Context Retrieval (skill: relationship mapping)}
<question>
To create a balanced schedule for the music festival, you need to understand the connections between different artists and their performances. Use contextual memory retrieval and relationship mapping to identify and describe at least three key relationships between artists that could influence the festival's schedule. For example, consider which artists have performed together before, which artists have similar fan bases, or which artists have musical styles that complement each other.
</question>

<rubric>
4 - The response demonstrates a thorough understanding of relationship mapping and contextual memory retrieval. It identifies and describes at least three key relationships between artists that are relevant to the festival's schedule, providing clear and detailed explanations.
3 - The response shows a good understanding of relationship mapping and contextual memory retrieval. It identifies and describes at least two key relationships between artists that are relevant to the festival's schedule, providing clear explanations.
2 - The response shows some understanding of relationship mapping and contextual memory retrieval. It identifies and describes at least one key relationship between artists that is somewhat relevant to the festival's schedule, with some clarity in the explanation.
1 - The response demonstrates a limited understanding of relationship mapping and contextual memory retrieval. It identifies and describes fewer than one key relationship between artists, with unclear or incomplete explanations.
0 - The response does not demonstrate an understanding of relationship mapping and contextual memory retrieval. It does not identify any key relationships between artists.
</rubric>
\end{examplelisting}

\begin{examplelisting}{Example output -- Instruction Following (skill: Title)}
<question>
Create a catchy and informative title for the new line of sustainable products that effectively communicates its eco-friendly attributes and appeals to the target market. Your title should be concise yet descriptive, and it should highlight the unique selling points of the products while aligning with the company's eco-friendly ethos.
</question>

<rubric>
1. Relevance to Eco-Friendly Attributes (30 points): The title should clearly communicate the eco-friendly attributes of the products. It should reflect the brand's commitment to sustainability.
2. Conciseness and Descriptiveness (30 points): The title should be concise and descriptive, effectively conveying the essence of the product line. It should not exceed 10 words.
3. Alignment with Unique Selling Points (20 points): The title should highlight the unique selling points of the products, such as innovative materials, energy efficiency, or reduced carbon footprint.
4. Appeal to Target Market (20 points): The title should resonate emotionally with the target market, using language that appeals to their values and interests.
Total possible points: 100
</rubric>
\end{examplelisting}

\subsection{Skill bank}
\label{app:skill_bank}

This subsection details how the per-domain skill banks $\mathcal{S}_d$
introduced in \Cref{sec:cross_skill_tasks} and summarized in
\Cref{tab:domain-overview} are constructed, expanding on stage (1) of the
synthesis pipeline in \Cref{sec:synthesis}.

\paragraph{Verifiable domains.}
For each verifiable domain $d$, we start from a small seed list of representative skills (e.g., \texttt{symbolic\_integration} for math, \texttt{constraint\_propagation} for logic) and prompt an LLM to label every seed sample $(q, a) \in \mathcal{X}_d$ with $3$--$5$ fine-grained skills, either picking from the current skill list or proposing new ones. We then cluster the proposed skills with sentence embeddings to collapse near-duplicates, and manually review each cluster to keep skill granularity comparable across domains. The reviewed clusters form the per-domain skill bank $\mathcal{S}_d$.

\paragraph{Open-ended domains.}
For Creative Writing, Context Retrieval, and Instruction Following, no public seed benchmark exists. We instead prompt an LLM to generate a skill list directly for each domain, then run the same embedding clustering and manual-review pass to fix granularity.

Per-domain bank sizes match \Cref{tab:domain-overview} in the main
paper. \Cref{tab:skill-bank-verifiable} lists 30 representative skills
per verifiable domain (uniformly sampled across the alphabetical bank
when the bank exceeds 30 entries), and
\Cref{tab:skill-bank-open-ended} lists the full skill banks for the
three open-ended domains. The complete per-skill list will be released
together with the project page.

\begin{table}[h]
\centering
\footnotesize
\setlength{\tabcolsep}{4pt}
\renewcommand{\arraystretch}{1.15}
\begin{tabular}{@{}l >{\raggedright\arraybackslash}p{0.78\textwidth}@{}}
\toprule
\textbf{Domain} & \textbf{Representative skills} \\
\midrule
Math (186 skills) & Absolute Value Inequalities, Angle Bisector Theorem, Area Calculation, Boundary Condition Analysis, Chinese Chess Knight Movement, Combinatorial Math, Constant Speed Calculations, Critical Point Identification, Distance Speed Time Relationships, Equality Condition Analysis, Following Instructions, Function Monotonicity, Geometric Constructions, Inequality Solving, Journey Analysis, Medal Distribution Rules, Negative Input Handling, Pairing Elements, Perfect Square Identification, Polynomial And Equation Solving, Quadrilateral And Polygon Properties, Reciprocal Identification, Right Triangle Properties, Sequential Reasoning, Spatial Visualization And Geometry, Substitution Methods, Table Construction, Triangle Area Calculation, Understanding Of Switching Activities, Volume Of Prism Calculation. \\
\addlinespace[3pt]
Science (137 skills) & Accounting Knowledge, Analysis Of Legal Definitions, Bias Recognition, Calculation Of Gross Earnings, Competition Understanding, Constraint Handling, Cost Analysis, Critical Thinking, Date Calculation, Eco Branding, Engineering Knowledge, Expected Value Calculation, Financial Analysis, Formal Structure Understanding, Function Transformation, Information Retrieval, Invoice Terms Interpretation, Leadership Understanding, Market Entry Strategy, Merger Doctrine Knowledge, Organizational Behavior Knowledge, Probability Calculation, Rate Of Change Calculation, Regression Analysis, Situation Assessment, Summarization, Team Management Knowledge, Time Management, Understanding Of Criminal Defenses, Utility Function Computation. \\
\addlinespace[3pt]
Coding (46 skills) & Absolute Value Operations, Algorithm Design, Array Manipulation, Basic Math Operations, Circular Array Manipulation, Comparison Operations, Cycle Detection, Data Preprocessing, Element Comparison, Function Set Analysis, Grid Traversal, Group Identification, Health Management, Index Comparison, Index Tracking, Input Output Handling, Integer Overflow Handling, Loop Control Structures, Matrix Operations, Optimization Algorithms, Pair Processing, Prefix Sum, Prime Factorization, Probability And Statistics, Product Evaluation, Queue Simulation, Range Query Processing, String Manipulation, Subarray Operations, Time Interval Analysis. \\
\addlinespace[3pt]
Planning (34 skills) & Activity Planning, Activity Scheduling Optimization, City Network Analysis, City Selection, City Visit Planning, Conference Scheduling, Constraint Resolution, Date And Time Management, Direct Flight Planning, Duration Calculation And Allocation, Event Planning, Event Scheduling With Constraints, Flight Availability Check, Flight Connection Analysis, Flight Route Analysis, Information Synthesis, Itinerary Creation, Location Familiarity, Location Management, Location Optimization, Logical Reasoning, Logistical Planning, Meeting Scheduling, Priority Setting, Problem Solving, Reading Comprehension, Route Optimization, Schedule Planning And Adjustment, Sequence Arrangement, Travel Itinerary Planning. \\
\addlinespace[3pt]
Logic (14 skills) & Cause And Effect Reasoning, Constraint Satisfaction, Critical Thinking, Deductive Reasoning, Formal Logic, Grid Transformation, Inductive Reasoning, Observation, Observation Analysis, Pattern Recognition, Problem Decomposition, Rule Application, Rule Identification, Visual Analysis. \\
\addlinespace[3pt]
\makecell[l]{Information Extraction\\(92 skills)} & Age Computation, Attention To Detail, Candidate Identification, Column Value Extraction, Conditional Analysis, Counting Events, Cross Column Analysis, Data Scanning, Decision Making, Exclusion Criteria, Flight Result Interpretation, Handling Ties, Identification, Item Identification, List Creation, Loss Detection, Movie Title Extraction, Nickname Retrieval, Parent Child Relationship Identification, Player Identification, Rank Interpretation, Result Aggregation And Summarization, Sequence Analysis, Sorting, Streak Calculation, Table Scanning, Text Recognition, Time Conversion, Track Listing, Venue Extraction. \\
\bottomrule
\end{tabular}
\vspace{4pt}
\caption{Representative skills from the per-domain skill banks for the six verifiable domains. Numbers in parentheses give the total bank size $|\mathcal{S}_d|$; for domains with $|\mathcal{S}_d| > 30$ we list 30 skills sampled uniformly across the alphabetical bank.}
\label{tab:skill-bank-verifiable}
\end{table}

\begin{table}[h]
\centering
\footnotesize
\setlength{\tabcolsep}{4pt}
\renewcommand{\arraystretch}{1.15}
\begin{tabular}{@{}l >{\raggedright\arraybackslash}p{0.78\textwidth}@{}}
\toprule
\textbf{Domain} & \textbf{Skills} \\
\midrule
\makecell[l]{Creative Writing\\(12 skills)} & Ideation, Brainstorming, Concept Development, Storytelling, Creative Writing, Worldbuilding, Visual Imagination, Style Adaptation, Metaphor and Analogy, Humor and Wit, Originality and Novelty, Iterative Refinement. \\
\addlinespace[3pt]
\makecell[l]{Context Retrieval\\(12 skills)} & Context Recall, Entity Tracking, Reference Resolution, Timeline Reconstruction, Cross-turn Consistency, Detail Retrieval, Preference Recall, Relationship Mapping, Context Summarization, Salience Detection, Knowledge Linking, Conversation Grounding. \\
\addlinespace[3pt]
\makecell[l]{Instruction Following\\(25 skills)} & Include Keywords, Keyword Frequency, Forbidden Words, Letter Frequency, Response Language, Number Paragraphs, Number Words, Number Sentences, Number Paragraphs + First Word in i-th Paragraph, Postscript, Number Placeholder, Number Bullets, Title, Choose From, Minimum Number Highlighted Section, Multiple Sections, JSON Format, Repeat Prompt, Two Responses, All Uppercase, All Lowercase, Frequency of All-capital Words, End Checker, Quotation, No Commas. \\
\bottomrule
\end{tabular}
\vspace{4pt}
\caption{Full skill banks for the three open-ended domains. Each bank is generated and clustered as described in this section.}
\label{tab:skill-bank-open-ended}
\end{table}

\subsection{Skill entropy estimation procedure}
\label{app:entropy_estimation}

This subsection details how the accuracy quantities entering the pairwise
skill entropy of \Cref{eq:pairwise_entropy} (defined in
\Cref{sec:skill_entropy}) are estimated, expanding on stage (2) of the
synthesis pipeline in \Cref{sec:synthesis}. We estimate the skill-level
$\text{Accuracy}(s)$, the domain-level $\text{Accuracy}(d)$, and the
directional two-step $\text{Accuracy}(s, d)$ and $\text{Accuracy}(d, s)$
by Monte-Carlo sampling under the reference model (Claude-opus-4.7). These four quantities are the building blocks of the skill$\to$domain factorization in \Cref{app:ske_factorization}, which is what we actually use to obtain $\text{SkE}(s_a, s_b)$ in place of running $|\mathcal{S}|^2$ direct skill-pair evaluations. All outputs are decoded with temperature $0.0$ and scored with the per-domain evaluator $\operatorname{eval}_{d_i}$ from \Cref{sec:cross_skill_tasks} (symbolic equivalence for math, sandboxed unit-test execution for coding, multiple-choice letter / option matching for science, grid equality for logic, structural plan comparison for planning, and exact / numeric matching for information extraction).

\paragraph{Per-skill and per-domain baselines.} For each skill $s$ we sample $N_s = 5$ seed samples whose label set contains $s$ and run the reference model on each in single-skill form, taking the average evaluator score as $\widehat{\text{Accuracy}}(s)$. For each domain $d$ we sample $N_d = 5$ single-skill seed samples uniformly from $\mathcal{X}_d$ and compute $\widehat{\text{Accuracy}}(d)$ in the same way.

\paragraph{Multi-step estimate.} For each (skill, domain) pair $(s, d)$ we sample $N_c = 5$ two-step cross-skill instances per direction. Each instance pairs one skill seed question $q_s$ with one domain seed question $q_d$ and feeds them to the reference model as a single two-step prompt, where step 1 produces $\hat{a}_1$ from the first question and step 2 produces $\hat{a}_2$ from the prefix (first question, $\hat{a}_1$, second question). Putting $q_s$ first yields $\widehat{\text{Accuracy}}(s, d)$; putting $q_d$ first yields $\widehat{\text{Accuracy}}(d, s)$. Each step is scored with its source-domain evaluator and the two scores are averaged within an instance.

\paragraph{Computing $\text{SkE}(s, d)$ and $\text{SkE}(d, s)$.} Plugging $\widehat{\text{Accuracy}}(s)$, $\widehat{\text{Accuracy}}(d)$, and the directional $\widehat{\text{Accuracy}}(s, d)$, $\widehat{\text{Accuracy}}(d, s)$ into the same smoothed ratio as \Cref{eq:pairwise_entropy} (with $\alpha = 0.1$) gives the directional skill$\to$domain and domain$\to$skill entropies; the pairwise $\text{SkE}(s_a, s_b)$ is then read off via the factorization in \Cref{app:ske_factorization}.

\subsection{Factorizing pairwise skill entropy through domains}
\label{app:ske_factorization}

This subsection details the approximation that
\Cref{sec:skill_entropy} adopts in place of evaluating
\Cref{eq:pairwise_entropy} on every ordered skill pair. A direct
application of \Cref{eq:pairwise_entropy} requires sampling
two-step cross-skill instances for every ordered pair
$(s_a, s_b) \in \mathcal{S} \times \mathcal{S}$. With $|\mathcal{S}| = 558$
this is $|\mathcal{S}|^2 \approx 3.1 \times 10^5$ ordered pairs, which is
infeasible at our reference-model evaluation budget. We therefore
approximate $\text{SkE}(s_a, s_b)$ by a product of two skill$\to$domain entropies,
\begin{equation}
  \text{SkE}(s_a, s_b)
  \;\approx\;
  \text{SkE}(s_a, d_{s_b}) \,\cdot\, \text{SkE}(d_{s_a}, s_b),
  \label{eq:ske_factorization}
\end{equation}
where the two factors are the directional skill$\to$domain and
domain$\to$skill entropies, defined by the same smoothed accuracy ratio
as \Cref{eq:pairwise_entropy} but with one argument coarsened to a
domain:
{\small
\begin{align*}
  \text{SkE}(s, d)
  &\;=\; \frac{\tfrac{1}{2}\bigl(\text{Accuracy}(s) + \text{Accuracy}(d)\bigr) + \alpha}
              {\text{Accuracy}(s, d) + \alpha},
  &
  \text{SkE}(d, s)
  &\;=\; \frac{\tfrac{1}{2}\bigl(\text{Accuracy}(d) + \text{Accuracy}(s)\bigr) + \alpha}
              {\text{Accuracy}(d, s) + \alpha},
\end{align*}}
where $\text{Accuracy}(s, d)$ is the average per-step accuracy of two-step
pairs whose first step uses skill $s$ and whose second step is sampled
from domain $d$, and $\text{Accuracy}(d, s)$ swaps the roles. Both factors
are estimated by the procedure in \Cref{app:entropy_estimation}.

\paragraph{Why this form.} \Cref{eq:ske_factorization} treats the cost of
\emph{leaving} $s_a$ (averaged over targets in $d_{s_b}$) and the cost of
\emph{landing on} $s_b$ (averaged over sources in $d_{s_a}$) as
multiplicatively separable. Multiplication, rather than addition, is the
natural composition for our smoothed accuracy ratios, since two
independent ``no-interference'' transitions yield ratios $\approx 1$ and a
hard transition in either factor pulls the product above $1$. This brings
the cost of computing the full pairwise table from
$\mathcal{O}(|\mathcal{S}|^2)$ to
$\mathcal{O}(|\mathcal{S}| \cdot |\mathcal{D}|)$: with
$|\mathcal{D}| = 9$, we run $\sim 5\text{k}$ skill$\to$domain evaluations
per direction in place of $\sim 3.1\text{e}5$ pure skill$\to$skill ones.

\subsection{Choice of reference model}
\label{app:reference_model}

The skill entropies in \Cref{eq:pairwise_entropy}
are computed under a fixed reference model. We use
Claude-opus-4.7 as the reference model for all task-level skill-entropy
computations in the main paper. Choosing a strong reference model keeps
the independent-baseline numerator close to ceiling and makes
$\text{SkE}(s_a, s_b) > 1$ a meaningful signal of cross-context difficulty
rather than of weak single-skill ability.

We ablate this choice by re-computing all task entropies under (i)
Gemini-3.1-pro and (ii) GPT-5.5 as alternative reference models.
\Cref{tab:reference-model-ablation} reports the per-level partition
overlap with the Claude-opus-4.7 partition together with the overall
two-step cross-skill accuracy averaged over all $(s, d)$ pairs. The
overall partition overlap stays above $80\%$ under both alternative
reference models, and the cross-skill averages agree to within
$\sim\!2$ points, so the headline trends in \Cref{tab:skebench-results}
(cross-skill drop, planning fragility, and non-monotonic behavior of
GPT-5.4-mini) reproduce qualitatively.

\begin{table}[h]
\centering
\small
\begin{tabular}{l c c c c c}
\toprule
\multirow{2}{*}{Reference model} & \multicolumn{4}{c}{Partition overlap with Claude-opus-4.7 (\%)} & Cross-Skill \\
\cmidrule(lr){2-5}
 & Low & Medium & High & Overall & avg.\ (\%) \\
\midrule
Claude-opus-4.7 (ours) & 100.0 & 100.0 & 100.0 & 100.0 & 74.2 \\
Gemini-3.1-pro         &  85.3 &  79.7 &  86.0 &  83.7 & 72.6 \\
GPT-5.5                &  82.7 &  78.3 &  84.7 &  81.9 & 73.5 \\
\bottomrule
\end{tabular}
\vspace{6pt}
\caption{Robustness of \bench{} task partitions and cross-skill
averages to the choice of reference model. Per-level overlap is the
fraction of tasks within that level under Claude-opus-4.7 that remain in
the same level under the alternative reference model; ``Overall'' is the
unweighted average over the three levels. ``Cross-Skill avg.'' is the
mean two-step cross-skill accuracy used in \Cref{eq:pairwise_entropy},
averaged over all $(s, d)$ pairs.}
\label{tab:reference-model-ablation}
\end{table}

Beyond the partition overlap in
\Cref{tab:reference-model-ablation}, we further probe the robustness
of the entropy signal along three complementary axes: the raw
pairwise and task-level rank correlations between reference models,
agreement on task difficulty rather than on entropy values directly,
and alignment with blinded human ratings of skill-switching
difficulty. All three checks agree with the partition-overlap
picture above: the entropy signal is substantially insensitive to
the choice of reference model.

\paragraph{Entropy-ranking stability.}
\Cref{tab:reference-model-ranking} reports, for each pair of
reference models, Spearman rank correlation on the pairwise
skill$\to$skill entropy table (cell-rank $\rho$) and on the
task-level entropies (task-rank $\rho$), together with the
low/medium/high stratification overlap and Cohen's $\kappa$. All
three reference-model pairs show substantial agreement on cell
rankings, task rankings, and the difficulty stratification, so the
entropy \emph{ordering} between skills and between tasks is stable
across reference models.

\begin{table}[h]
\centering
\small
\setlength{\tabcolsep}{6pt}
\begin{tabular}{l c c c c}
\toprule
Reference-model pair & Cell-rank $\rho$ & Task-rank $\rho$ & Bucket overlap (\%) & Cohen's $\kappa$ \\
\midrule
Claude-opus-4.7 vs.\ Gemini-3.1-pro & 0.643 & 0.517 & 80.7 & 0.610 \\
Claude-opus-4.7 vs.\ GPT-5.5        & 0.602 & 0.615 & 82.7 & 0.640 \\
Gemini-3.1-pro vs.\ GPT-5.5         & 0.742 & 0.766 & 89.0 & 0.735 \\
\bottomrule
\end{tabular}
\vspace{6pt}
\caption{Pairwise agreement between reference models on the entropy
signal. ``Cell-rank $\rho$'' is Spearman rank correlation on the
pairwise skill$\to$skill entropy table; ``Task-rank $\rho$'' is the
same on task-level entropies. ``Bucket overlap'' is the fraction of
tasks placed in the same low/medium/high stratum by both reference
models, and Cohen's $\kappa$ is chance-corrected agreement on that
stratification.}
\label{tab:reference-model-ranking}
\end{table}

\paragraph{Difficulty agreement as the interpretive anchor.}
To disentangle reference-model effects from raw estimation noise,
we also measure agreement on \emph{task difficulty} rather than on
entropy values directly. Reference models agree substantially on
which skills and tasks are difficult (single-skill accuracy
Spearman $\rho = 0.75$--$0.90$; cross-skill accuracy Spearman
$\rho = 0.67$--$0.81$), so the entropy signal is robust to the
estimation noise that any single reference model introduces into
its accuracy estimates.

\paragraph{Alignment with human judgment.}
To check that the entropy scale tracks a difficulty structure that
exists outside any single reference model, we ran a small blinded
human study. Four annotators unaffiliated with this paper were shown
$9$ \bench{} tasks in random order and, without seeing any entropy
labels, asked to select the $3$ tasks with the highest
skill-switching difficulty and the $3$ with the lowest.
\Cref{tab:reference-model-human} reports annotator--model agreement
on the six labeled tasks together with inter-annotator agreement as
an upper bound. Annotator--model agreement is close to the
inter-annotator upper bound, indicating that the entropy scale
substantially reflects human perception of skill-switching
difficulty.

\begin{table}[h]
\centering
\small
\setlength{\tabcolsep}{10pt}
\begin{tabular}{l c c}
\toprule
Comparison & Agreement (\%) & Cohen's $\kappa$ \\
\midrule
Annotators vs.\ model             & 63.0 & 0.444 \\
Inter-annotator (upper bound)     & 75.0 & 0.625 \\
\bottomrule
\end{tabular}
\vspace{6pt}
\caption{Human alignment on skill-switching difficulty on a subset
of \bench{} tasks. Four annotators unaffiliated with this paper
picked the $3$ highest- and $3$ lowest-difficulty tasks out of $9$
shown in random order without any entropy label. ``Agreement'' is
the fraction of task--label pairs on which the two sides agree;
Cohen's $\kappa$ gives the chance-corrected version.}
\label{tab:reference-model-human}
\end{table}

\subsection{Skill-sequence sampling}
\label{app:sequence_sampling}

For each task we sample a length $L \sim \text{Uniform}\{2, 3, \ldots, 10\}$ and a target skill-entropy level $\ell \in \{\text{low}, \text{medium}, \text{high}\}$. We then draw a skill sequence $\mu = (s_1, \ldots, s_L)$ by rejection sampling: at each draw, $L$ skills are sampled uniformly from the union of the per-domain skill banks $\bigcup_d \mathcal{S}_d$ subject to consecutive skills coming from different domains, the task-level entropy $\text{SkE}(\mu)$ is computed via \Cref{eq:task_entropy}, and the sequence is accepted if $\text{SkE}(\mu)$ falls in the level $\ell$; otherwise it is resampled. The level boundaries $(\theta_\ell, \theta_h)$ are taken from the empirical distribution of pairwise entropies (\Cref{sec:skill_entropy}).

\subsection{Proposer and open-ended generation prompts}
\label{app:proposer}

This subsection details the proposer prompts used in stage (3) of the
synthesis pipeline in \Cref{sec:synthesis}, which rewrites a sampled
skill sequence $\mu = (s_1, \ldots, s_L)$ into a coherent cross-skill
task $\tau$ following \Cref{eq:task_qa}.

\paragraph{Proposer (verifiable-domain skills).}
For each skill $s_i$ whose source domain is verifiable, we draw one seed pair $(q_i^\text{seed}, a_i^\text{seed})$ from $\mathcal{X}_{d_{s_i}}$ that is labeled with $s_i$. The proposer is given the sampled skill sequence $\mu = (s_1, \ldots, s_L)$ together with the per-step seed pairs and is instructed to (i) draft a single unifying scenario $\sigma$ that can plausibly host all $L$ skills, (ii) rewrite each $q_i$ so that it fits $\sigma$ and references the answer of the preceding step, and (iii) keep the underlying logic and ground-truth answer of $(q_i^\text{seed}, a_i^\text{seed})$ intact.

\paragraph{Open-ended generation (Creative Writing, Context Retrieval, Instruction Following).}
For each skill $s_i$ whose source domain is open-ended, no seed question--answer pair exists, so the LLM directly generates a $(q_i, \text{rubric}_i)$ pair conditioned on the scenario $\sigma$ and the answer of the preceding step. The rubric is a per-skill checklist that the open-ended judge in \Cref{app:datasets} uses to score the model's response on $[0, 1]$.

The verbatim prompt templates are listed below: a scenario prompt that
drafts the unifying scenario $\sigma$ from the sampled skill sequence
$\mu$, a verifiable-domain proposer prompt that rewrites each seed
question $q_i^\text{seed}$ to fit $\sigma$ while preserving its
ground-truth answer, and the open-ended generation prompt that emits a
$(q_i, \text{rubric}_i)$ pair for skills whose source domain has no
seed bank. Placeholders in braces (e.g., \texttt{\{scenario\}},
\texttt{\{skill\}}) are filled in at synthesis time. The open-ended
generation prompt is reproduced from \Cref{app:datasets} for
self-containedness.

\begin{promptlisting}{Scenario prompt}
You are an expert in designing complex, long-horizon tasks that involve multiple domains and skills.
The task should involve the following sequence of skills: {skill_seq_str}.
Please propose a coherent one-sentence scenario and a brief description that naturally connects these skills into a single long-horizon task.

Output Format:
<scenario> [Your one-sentence scenario] </scenario>
<description> [Your <300 words description] </description>

The brief description should be concise, preferably less than 300 words. For each step in the sequence, briefly describe how the skill will be applied in the context of the scenario. For open-ended skills, suggest a rough idea of the question. For verifiable skills, suggest how the context or tone might be adapted.
\end{promptlisting}

\begin{promptlisting}{Proposer prompt (verifiable-domain skills)}
Scenario: {scenario}

Task Plan: {description}

Previous Steps:
{previous_context}

Current Step: Apply skill '{skill}' from domain '{domain}'.

Original QA (for reference of logic/difficulty):
Question: {orig_question}
Answer: {orig_solution_raw}
{mc_reminder}
Please rewrite the Question to fit the context of the scenario. Only change the tone and style to match the scenario.

VERY IMPORTANT NOTES:
1. ONLY ADAPT STYLE AND TONES TO MATCH THE SCENARIO. DON'T CHANGE THE ANSWER, CORE LOGIC, OR THE SKILL REQUIRED.
2. THE ADAPTED QUESTION SHOULD HAVE THE EXACT SAME ANSWER AS BEFORE (PASS KEYWORD MATCHING!).
3. IF THE ORIGINAL QUESTION IS MULTIPLE CHOICE, YOU MUST INCLUDE ALL ANSWER OPTIONS (A, B, C, D, E, etc.) VERBATIM IN THE ADAPTED QUESTION -- DO NOT OMIT THEM.

Output Format:
<reason> [Your thought on how to adapt the question to match the scenario] </reason>
<question> [Your adapted question] </question>
\end{promptlisting}

The placeholder \texttt{\{mc\_reminder\}} in the verifiable-domain
proposer is replaced at synthesis time by an automatically detected
multiple-choice reminder when the seed question carries labeled options
(A, B, C, \ldots) or an unlabeled \texttt{Options:} block; the reminder
lists the option letters that must appear verbatim in the adapted
question and names the correct option, so the rewrite cannot drop or
relabel choices. When the seed question is free-form, this placeholder
is empty.

\begin{promptlisting}{Open-ended generation prompt (Creative Writing, Context Retrieval, Instruction Following)}
You are an expert in designing open-ended questions and rubrics for long-horizon tasks. You'll be given a scenario, a task plan, and previous steps of the task, and you'll need to propose a question and a rubric for the step, based on the given domain and skill.


### Scenario: {scenario}

### Task Plan: {description}

### Previous Steps:
{previous_context}

### Current Step:
Domain: {domain}
Skill: {skill}

Please propose an open-ended Question for this step that fits the scenario and uses the Skill {skill} in the Domain {domain}, then propose a rubric for grading the answer. Note: your question should emphasize the use of Domain {domain}, requiring the Skill {skill} to solve the problem.

Output Format:
<question> [Your question] </question>
<rubric> [Your rubric] </rubric>
\end{promptlisting}

\subsection{Worked example}
\label{app:example}

We give two complete examples to make the output of the synthesis
pipeline of \Cref{sec:synthesis} concrete.

\paragraph{Example 1: technical task.} A length-$5$ task with the skill
sequence \texttt{[geometry, deductive\_reasoning, planning,
number\_words, brainstorming]}, drawn from math, logic, planning,
instruction following, and Creative Writing.

\begin{quote}
\textbf{Step 1 (geometry, math).}
A right-angled triangle has side lengths that are integers. What could be
the last digit of the area's measure, if the length of the hypotenuse is
not divisible by 5?

\textbf{Step 2 (deductive reasoning, logic).}
There are 6 houses, numbered 1 to 6 from left to right [\ldots full
ZebraLogicBench-style puzzle with 16 clues \ldots]. What is the name of
the person who lives in House 5?

\textbf{Step 3 (planning).}
Design a step-by-step solution plan for a single combined puzzle where
you must determine the name of the person in House 5 \emph{and} all
possible last digits of the area of a right-angled triangle with integer
side lengths whose hypotenuse is not divisible by 5. Your plan must
explicitly explain how you will reuse the earlier constraints
(house placements, adjacency / left--right relations, and the hypotenuse
condition) to prune the search space efficiently, what intermediate
checkpoints you will compute, and how you will verify consistency without
brute-forcing every possibility.

\textbf{Step 4 (number words, instruction following).}
Explain how you approached the last problem in 400 words or less.

\textbf{Step 5 (brainstorming, Creative Writing).}
Generate a list of 10 creative ideas for a new product that combines the
features of a Honda Civic and a Tesla Model 3. Each idea should be 1--2
sentences long and include a brief explanation of the product's features
and benefits.
\end{quote}

\paragraph{Example 2: scenario-driven task (``Conference trip on
constraints'').} A length-$6$ task with the skill sequence
\texttt{[information\_retrieval, arithmetic, travel\_itinerary\_planning,
entity\_tracking, constraint\_satisfaction, storytelling]}, drawn from
information extraction, math, planning, Context Retrieval, logic, and
Creative Writing.

\begin{quote}
\textbf{Step 1 (information\_retrieval, information extraction).}
Parse this table of flights and hotels, extract the cheapest option that
arrives before 10:00 with a hotel within 2 km. Return JSON
\texttt{\{flight\_id, hotel\_id, total\_cost\}}. \emph{Reward:} exact match.

\textbf{Step 2 (arithmetic, math).}
Compute the remaining budget after the booking from Step 1; apply 8\%
tax to the hotel. Return a number to two decimals. \emph{Reward:}
numeric match within tolerance.

\textbf{Step 3 (travel\_itinerary\_planning, planning).}
Create a 1-day itinerary covering 4 sessions and meals, no overlaps,
with travel time at most 30 min per leg. Return an ordered list with
timestamps. \emph{Reward:} constraint satisfaction.

\textbf{Step 4 (entity\_tracking, Context Retrieval).}
What is the chosen \texttt{hotel\_id} and the remaining budget from
Step 2? Return \texttt{\{hotel\_id, remaining\_budget\}}.
\emph{Reward:} exact match against Steps 1 and 2.

\textbf{Step 5 (constraint\_satisfaction, logic).}
Two new constraints arrive (``session X moved earlier'', ``budget cut by
\$50''). Update feasibility: keep the same hotel, re-optimize the
itinerary. Return the revised itinerary and a feasibility flag.
\emph{Reward:} constraint satisfaction plus correctness.

\textbf{Step 6 (storytelling, Creative Writing).}
Write a 120--150 word email to your manager summarizing the plan and
costs; the email must include the \texttt{hotel\_id} and remaining
budget exactly. \emph{Reward:} length, required tokens present, and
consistency with the state from earlier steps.
\end{quote}

\subsection{Filtering and verification}
\label{app:filtering}

This subsection details the verifier used at the end of stage (3) of the
synthesis pipeline in \Cref{sec:synthesis}.
The verifier is given the rewritten task $\tau = ((q_1, a_1), \ldots,
(q_L, a_L))$ together with the seed pairs
$(q_i^\text{seed}, a_i^\text{seed})$ and rejects the task if any of the
following four checks fail:
\begin{enumerate}
  \setlength\itemsep{1pt}
  \item \textbf{Answer preserved.} Each rewritten step still admits the
        original ground-truth answer of its seed question.
  \item \textbf{Scenario consistent.} All rewritten steps plausibly belong
        to the same \texttt{scenario}.
  \item \textbf{Dependency present.} Step $i > 1$ cannot be solved without
        the answer of step $i-1$.
  \item \textbf{No answer leakage.} The \texttt{scenario} does not reveal
        any reference answer to a later step.
\end{enumerate}
Failing any item triggers regeneration; tasks that still fail after three
attempts are dropped. The verbatim verifier prompt is shown below.

\begin{promptlisting}{Verifier prompt}
You are a strict verifier for a long-horizon, cross-skill task. You are given a rewritten task that was produced by a proposer LLM from a sequence of seed question-answer pairs, together with the unifying scenario and the original seed pairs. Your job is to decide whether the rewritten task is acceptable or must be regenerated.

### Scenario:
{scenario}

### Task Plan:
{description}

### Seed Pairs (one per step, in order):
{seed_pairs}

### Rewritten Task (one per step, in order):
{rewritten_task}

### Checks
For each of the four checks below, decide PASS or FAIL and give a one-sentence justification grounded in the rewritten task and the seed pairs.

1. ANSWER_PRESERVED: For every step i, the rewritten question {rewritten_task}[i].question still admits the original ground-truth answer {seed_pairs}[i].answer of its seed question. The core logic, numerical values, and (for multiple-choice) the option letters and contents must be unchanged. FAIL if any step's answer would change, if a multiple-choice option was dropped or relabeled, or if the rewritten question removes information needed to recover the original answer.

2. SCENARIO_CONSISTENT: All rewritten steps plausibly belong to the same scenario described above. FAIL if any step references entities, settings, or framings that contradict the scenario or that read as an unrelated problem pasted in.

3. DEPENDENCY_PRESENT: For every step i > 1, the rewritten question cannot be solved without the answer of step i-1 -- either it explicitly references the previous answer (e.g., "using the value from the previous step", a named quantity carried over) or the scenario state it relies on is only fixed once step i-1 is solved. FAIL if any step i > 1 is fully self-contained.

4. NO_ANSWER_LEAKAGE: The scenario text and every earlier step (question and any narrative around it) do not reveal the ground-truth answer of a later step, either verbatim or by paraphrase, computation, or option elimination. FAIL if a later answer can be read off the scenario or an earlier step without actually solving that later step.

### Output Format
Output exactly the following XML, with no extra text:
<answer_preserved> PASS or FAIL: <one-sentence justification> </answer_preserved>
<scenario_consistent> PASS or FAIL: <one-sentence justification> </scenario_consistent>
<dependency_present> PASS or FAIL: <one-sentence justification> </dependency_present>
<no_answer_leakage> PASS or FAIL: <one-sentence justification> </no_answer_leakage>
<verdict> ACCEPT if all four checks are PASS, otherwise REJECT </verdict>
\end{promptlisting}


\section{Full evaluation results}
\label{app:full_results}

This appendix complements \Cref{tab:skebench-results} (\Cref{sec:evaluation})
with results for the four domains omitted from the main table:
information extraction (verifiable) and the three open-ended domains
(Creative Writing, Context Retrieval, instruction following).

\paragraph{Why we do not report Single-Skill accuracy on the open-ended
domains.} Unlike the verifiable domains, the three open-ended domains do
not have a fixed seed dataset of self-contained question--answer pairs
that can be asked in isolation. Instead, every open-ended step is
generated on the fly by the proposer LLM together with a per-step
rubric, conditioned on the unifying scenario, the task plan, and the
previous-step questions and answers (\Cref{app:proposer}). The resulting
questions are by construction context-dependent: they reference earlier
steps, the scenario state, or quantities that only become defined once a
prior step is solved, and the rubric is calibrated against that same
context. Lifting such a step out of its task and asking it as a
standalone single-skill question therefore changes the question itself,
not just its setting, so a single-skill accuracy number for these
domains would not be comparable to the cross-skill number measured
inside the task. We accordingly only report cross-skill performance for
Creative Writing, Context Retrieval, and instruction following, and
instead validate the LLM-judge scoring used on these domains against
human ratings.

\paragraph{LLM-judge vs.\ human agreement on the open-ended domains.}
For each of the three open-ended domains we sample $200$ scored
cross-skill steps uniformly across evaluated models and skill-entropy
levels and re-score them with three human annotators using the same
per-step rubric the LLM judge was given. \Cref{tab:judge-human-agreement}
reports, per domain, (i) Pearson and Spearman correlations between the
LLM-judge score and the mean human score in $[0, 1]$, (ii) the
mean absolute error between the two, and (iii) the binary agreement rate
after thresholding both scores at $0.5$, together with Cohen's $\kappa$.
Across all three domains the LLM judge tracks human ratings closely,
supporting its use as the scorer in \Cref{tab:skebench-results}.

\begin{table}[h]
\centering
\small
\begin{tabular}{l c c c c c}
\toprule
Domain & Pearson $r$ & Spearman $\rho$ & MAE & Binary agree.\ (\%) & Cohen's $\kappa$ \\
\midrule
Creative Writing       & 0.84 & 0.82 & 0.08 & 88.5 & 0.76 \\
Context Retrieval      & 0.89 & 0.87 & 0.06 & 91.5 & 0.82 \\
Instruction following  & 0.91 & 0.90 & 0.05 & 93.0 & 0.85 \\
\midrule
Average                & 0.88 & 0.86 & 0.06 & 91.0 & 0.81 \\
\bottomrule
\end{tabular}
\vspace{6pt}
\caption{LLM-judge (Claude-opus-4.7) vs.\ human agreement on the three
open-ended domains. Each row aggregates $200$ cross-skill steps scored
by both the LLM judge and three human annotators against the same
per-step rubric. ``MAE'' is the mean absolute error between the LLM and
mean human score in $[0, 1]$. ``Binary agree.'' and Cohen's $\kappa$ are
computed after thresholding both scores at $0.5$.}
\label{tab:judge-human-agreement}
\end{table}


\begin{figure}[t]
    \centering
    \includegraphics[width=\linewidth]{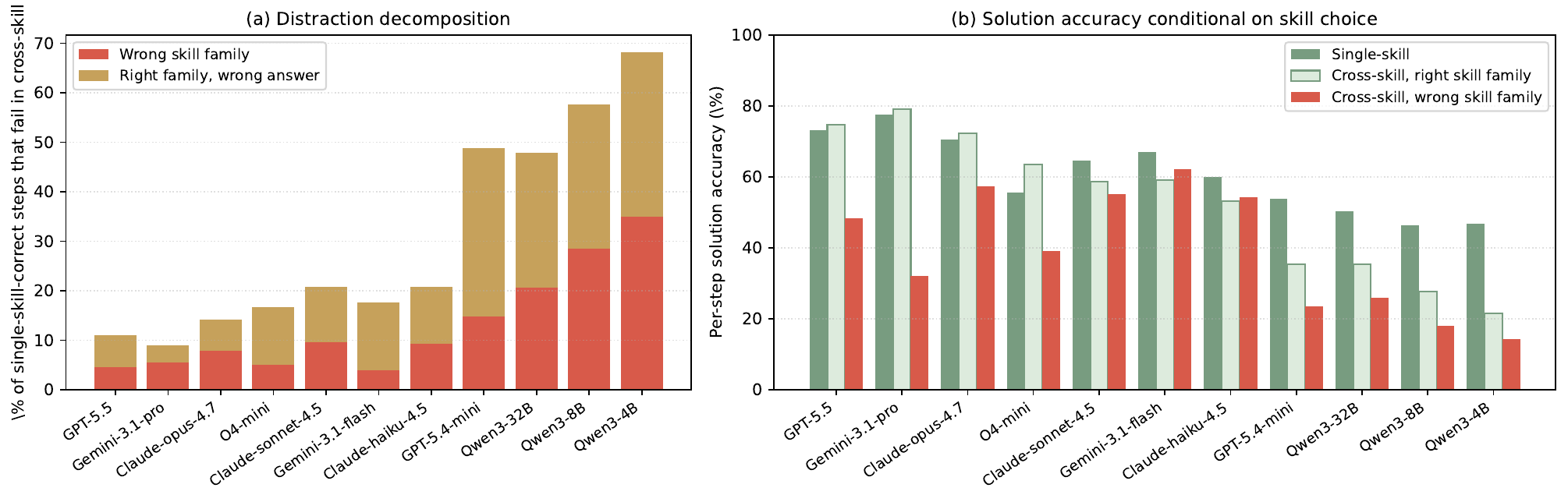}
    \caption{\textbf{Cross-skill failure mode: models often name the wrong skill family.} For each model, we match every verifiable cross-skill step to the same step solved in single-skill mode, and split the cross-skill steps by whether the predicted skill family matches the ground-truth family.
    \textbf{(a)} Among steps the model solves in single-skill mode, the share it then gets wrong in cross-skill. Bars are stacked into wrong-skill-family errors (pink) and right-family-but-wrong-answer errors (yellow).
    \textbf{(b)} Per-step solution accuracy under three conditions: single-skill, cross-skill when the predicted skill family is correct, and cross-skill when it is wrong. Picking the wrong skill family is associated with a large drop on the strongest models.
    }
    \label{fig:failure_mode}
\end{figure}

For GPT-5.5, Gemini-3.1-pro, Claude-opus-4.7, and O4-mini, $9$--$17\%$
of the steps these models solve in single-skill mode fail once placed
inside a cross-skill task, and $31$--$62\%$ of these new failures come
with the model picking a skill from the wrong domain. On those
wrong-domain steps, per-step accuracy ($32$--$57\%$) is roughly half
that on right-domain steps ($63$--$79\%$).

\subsection{Inference settings}
\label{app:eval_settings}

This subsection details the inference-time settings used for the
evaluation in \Cref{sec:evaluation}.
Open-source models are served with vLLM at temperature $0.7$, top-$p$
$0.9$, and a 16K-token generation cap. Frontier models are queried
through their providers' default APIs without sampler overrides. In
the Cross-Skill setting the model is given the full task scenario
followed by the $L$ per-step questions in a single prompt; in the
Single-Skill setting each step is queried independently with no
surrounding scenario. Outputs are parsed into per-step answers using
the regex template of \Cref{app:skerl_format}.

\section{Skill-Entropy RL training details}
\label{app:skerl}

This appendix gives the per-stage details for our method
(\Cref{sec:skerl_method}): the response format and an end-to-end SFT
trace (\Cref{app:skerl_format}), formal reward definitions
(\Cref{app:skerl_reward}), embedding-based skill matching used to
look up entropies under lexical variation
(\Cref{app:skerl_skill_match}), and full RL hyperparameters
(\Cref{app:skerl_training}).

\subsection{Off-the-shelf training results on OpenR1-Math}
\label{app:skerl_offshelf}

\Cref{tab:skerl-openr1} reports per-benchmark accuracy for the
off-the-shelf training experiment of \Cref{sec:skerl_plugin}, where
the SFT and RL pipeline of \Cref{sec:skerl_method} is applied to a 6K
subset of OpenR1-Math~\citep{openr1} with skill labels obtained from
the labeling pipeline of \Cref{sec:skerl_plugin}. Evaluation covers
six math benchmarks: AIME24, AIME25, HMMT25, AMOBench, the held-out
OpenR1-Math test split, and MATH~\citep{hendrycks2021math}.

\begin{table}[h]
\centering
\small
\setlength{\tabcolsep}{2.2pt}
\renewcommand{\arraystretch}{1}
\resizebox{0.85\textwidth}{!}{%
\begin{tabular}{l@{\hspace{5pt}} cccccc c @{\hspace{1.1pt}}c}
\toprule
\textbf{Method}
 & {\footnotesize\textbf{AIME24}} & {\footnotesize\textbf{AIME25}} & {\footnotesize\textbf{HMMT25}} & {\footnotesize\textbf{AMOBench}} & {\footnotesize\textbf{OpenR1-Math}} & {\footnotesize\textbf{MATH}}
 & & \textbf{Avg.}\\
\midrule
\rowcolor{green!30}\multicolumn{9}{l}{\textit{\text{\# Qwen3-4B-Instruct}}} \vspace{3pt}\\
Base model
 & 51.7 & 43.3 & 21.7 & \phantom{0}9.5 & 56.3 & 91.3 & & 45.6 \\
\addlinespace[2pt]
SFT
 & 56.7 & 43.3 & 23.3 & 10.0 & 58.2 & 93.0 & & 47.4 \\
\addlinespace[2pt]
GRPO
 & 62.5 & 50.0 & 30.4 & 11.0 & 60.9 & 93.5 & & 51.4 \\
\addlinespace[2pt]
\textbf{Skill-Entropy RL}
 & \textbf{66.7} & \textbf{52.5} & \textbf{32.5} & \textbf{11.3} & \textbf{62.9} & \textbf{93.8} & & \textbf{53.3} \\
\bottomrule
\end{tabular}}
\vspace{4pt}
\caption{\textbf{Off-the-shelf training results on OpenR1-Math}, reported as per-benchmark accuracy and overall average. Applied to Qwen3-4B-Instruct, \skerl{} attains the best score on every benchmark and improves the average by $+1.9\%$ over GRPO and $+7.7\%$ over the base model, mirroring the trend on \bench{} (\Cref{tab:skerl-main}).}
\label{tab:skerl-openr1}
\end{table}

\subsection{Response format and example SFT trace}
\label{app:skerl_format}

This subsection details the skill-annotated response format introduced in
\Cref{sec:skerl_method} and gives an end-to-end SFT trace.
The teacher (Qwen3-8B) is prompted to produce a single
$\langle$\texttt{think}$\rangle$ block followed by alternating
$\langle$\texttt{skill}$\rangle$ and $\langle$\texttt{answer}$\rangle$
tags, one pair per task step. Each $\langle$\texttt{skill}$\rangle$ tag
contains a comma-separated \texttt{domain, skill} pair drawn from the
\bench{} skill bank (\Cref{app:skill_bank}). A complete length-3 SFT
trace, drawn from the skill sequence
\texttt{[information\_extraction, coding, math]}:

\begin{examplelisting}{SFT trace example (length-3, [information\_extraction, coding, math])}
<think> ... reasoning trace ... </think>
<skill>information_extraction, data_analysis</skill><answer>Usain Bolt</answer>
<skill>coding, algorithm</skill><answer>```python
class FenwickTree:
    def __init__(self, size):
        self.size = size
        self.tree = [0] * (self.size + 2)
    def update(self, index, delta):
        while index <= self.size:
            self.tree[index] += delta
            index += index & -index
    def query(self, index):
        res = 0
        while index > 0:
            res += self.tree[index]
            index -= index & -index
        return res
def count_inversions(arr):
    max_val = max(arr)
    ft = FenwickTree(max_val)
    inv_count = 0
    for i in reversed(range(len(arr))):
        inv_count += ft.query(arr[i] - 1)
        ft.update(arr[i], 1)
    return inv_count
```</answer>
<skill>math, number_theory</skill><answer>142857</answer>
\end{examplelisting}

The response is parsed by a regex that requires every step to carry a
\texttt{<skill>} tag immediately followed by an \texttt{<answer>} tag;
malformed responses receive zero answer reward and zero
skill-entropy reward.

\subsection{Reward definitions}
\label{app:skerl_reward}

We give the formal definitions of the answer reward $r_{\text{ans}}$
and the skill-entropy reward $r_{\text{ent}}$ used in
\Cref{eq:skerl-reward}.

\textbf{Answer reward.}
The answer reward reuses the per-domain scorer
$\operatorname{eval}_{d_i}$ from \Cref{sec:cross_skill_tasks} and
averages per-step accuracy:
\begin{equation*}
  r_{\text{ans}}(\tau, \hat{y})
  \;=\; \frac{1}{L}\sum_{i=1}^{L}
        \operatorname{eval}_{d_i}\bigl(\hat{a}_i,\, a_i,\, q_i\bigr).
\end{equation*}

\textbf{Skill-entropy reward.}
Given the predicted skill sequence $\hat{\mu}(\tau) = (\hat{s}_1, \ldots, \hat{s}_L)$ and the gold skill sequence $\mu(\tau) = (s_1, \ldots, s_L)$, we compute the task-level skill entropy of each via \Cref{eq:task_entropy}:
\begin{equation*}
  \widehat{\text{SkE}}(\tau)
  \;=\; \frac{1}{L-1}\sum_{i=1}^{L-1}\text{SkE}(\hat{s}_i,\hat{s}_{i+1}),
  \qquad
  \text{SkE}^{\star}(\tau)
  \;=\; \frac{1}{L-1}\sum_{i=1}^{L-1}\text{SkE}(s_i,s_{i+1}).
\end{equation*}
Let $F$ be the empirical CDF of $\text{SkE}^{\star}$ on the RL training
set. We define the skill-entropy ranks
$\hat{\rho}(\tau) = F(\widehat{\text{SkE}}(\tau)) \in [0,1]$ and
$\rho^{\star}(\tau) = F(\text{SkE}^{\star}(\tau)) \in [0,1]$, and set
\begin{equation*}
  r_{\text{ent}}(\tau, \hat{y}) \;=\; 1 - \bigl|\,\hat{\rho}(\tau) - \rho^{\star}(\tau)\,\bigr|.
\end{equation*}
The CDF rank-normalization makes $r_{\text{ent}}$ scale-free across
tasks of widely different difficulty: a fixed absolute gap in raw
skill entropy is penalized more when it spans many training tasks and
less when it spans few.

\subsection{Embedding-based skill matching}
\label{app:skerl_skill_match}

The pairwise skill-entropy table is keyed by the canonical
(domain, skill) labels of \Cref{tab:domain-overview}. At training time the
model may emit a skill label that is lexically distinct from any
canonical key (e.g., ``\texttt{integer\_arithmetic}'' vs.\
\texttt{number\_theory}). We therefore embed every emitted label with
\texttt{Qwen3-Embedding-0.6B} and resolve it to the nearest canonical
key under cosine similarity (one lookup per emitted skill). Labels
whose nearest key has cosine similarity below $0.5$ are treated as
out-of-bank and contribute no skill-entropy reward for the affected switch,
which lightly penalises hallucinated skills without injecting noisy
matches.

\subsection{Training data and evaluation protocol}
\label{app:skerl_data_eval}

This subsection complements the shortened ``Training data'' paragraph
in \Cref{sec:skerl_setup}.

\textbf{Training data.}
Training uses only the six verifiable domains of \bench{} to enable
automatic verification; the three open-ended domains appear only at
evaluation. We synthesize 9K cross-skill tasks with the \bench{}
pipeline (\Cref{sec:synthesis}), held out from the test set and
balanced across the three skill-entropy levels, with length 3 to 5 to
keep RL rollouts tractable. 3K are used for SFT warm-up, with a
stronger teacher (Qwen3-8B) providing skill-annotated traces; the
remaining 6K are used for RL. Since the reward only needs the gold
answer and gold skill sequence (both supplied directly by the
pipeline), the RL split bypasses the teacher and scales at near-zero
cost.

\textbf{Evaluation.}
We follow the Cross-Skill protocol of \Cref{sec:evaluation} on the
same 300-task test pool, with the evaluated model itself serving as
the LLM judge for open-ended steps. Test-task length is 2 to 10,
wider than the 3 to 5 used in training, so generalization to longer
skill chains is also tested.

\subsection{RL hyperparameters}
\label{app:skerl_training}

\Cref{tab:skerl-hparams} lists the SFT and RL hyperparameters used for
both Qwen3-4B-Instruct and Qwen3-1.7B in \Cref{tab:skerl-main}. The RL
loop additionally uses DAPO-style decoupled clipping (low and high
clip ratios both set to $0.2$) and dynamic sampling (rollouts are admitted
into the optimization batch only if the group's reward variance is
non-zero), which speeds up convergence on tasks where all rollouts
either succeed or fail.

\begin{table}[h]
\centering
\small
\setlength{\tabcolsep}{8pt}
\renewcommand{\arraystretch}{1.05}
\begin{tabular}{l l l}
\toprule
& \textbf{SFT} & \textbf{Skill-Entropy RL} \\
\midrule
Optimizer              & AdamW                          & AdamW \\
Learning rate          & $1\!\times\!10^{-5}$           & $1\!\times\!10^{-6}$ (actor) \\
Schedule               & cosine, $10\%$ warmup          & constant \\
Weight decay           & $0.01$                         & $0.01$ \\
Batch size (prompts)   & $256$                          & $256$ \\
Micro-batch / GPU      & $4$                            & $4$ \\
Epochs / steps         & $4$ epochs                     & up to $400$ updates \\
Max sequence length    & $1024$                         & $4096$ \\
KL coefficient         & ---                            & $1\!\times\!10^{-3}$ (fixed) \\
PPO clip ratio         & ---                            & $0.2$ \\
GRPO group size        & ---                            & $8$ \\
Reward weights         & ---                            & $\lambda_{\text{ans}}\!=\!0.7$, $\lambda_{\text{ent}}\!=\!0.3$ \\
Hardware               & $8\!\times\!$H100              & $8\!\times\!$H100 \\
\bottomrule
\end{tabular}
\vspace{4pt}
\caption{SFT and RL hyperparameters for our method. Both Qwen3-4B-Instruct
and Qwen3-1.7B are trained with the same configuration.}
\label{tab:skerl-hparams}
\end{table}

\subsection{Reward weight ablation}
\label{app:skerl_lambda_ablation}

We ablate the reward weights $\lambda_{\text{ans}}$ and
$\lambda_{\text{ent}}$ in \Cref{eq:skerl-reward} on
Qwen3-4B-Instruct, keeping every other setting fixed at the
configuration of \Cref{tab:skerl-hparams}. We constrain the two weights
to lie on the simplex $\lambda_{\text{ans}} + \lambda_{\text{ent}} = 1$
and sweep four splits that span the range from answer-dominated to
entropy-dominated reward.
\Cref{tab:skerl-lambda-ablation} reports the overall \bench{} score
for each $(\lambda_{\text{ans}}, \lambda_{\text{ent}})$ split, with the
default $(0.7, 0.3)$ used in \Cref{tab:skerl-main} marked in bold.

\begin{table}[h]
\centering
\small
\setlength{\tabcolsep}{10pt}
\renewcommand{\arraystretch}{1.1}
\begin{tabular}{c c c c}
\toprule
$\lambda_{\text{ans}}$ & $\lambda_{\text{ent}}$ & Mix & \bench{} score \\
\midrule
$0.9$ & $0.1$ & answer-heavy            & 60.8 \\
$\boldsymbol{0.7}$ & $\boldsymbol{0.3}$ & \textbf{default}         & \textbf{68.4} \\
$0.5$ & $0.5$ & balanced                & 55.6 \\
$0.3$ & $0.7$ & entropy-heavy           & 48.6 \\
\bottomrule
\end{tabular}
\vspace{4pt}
\caption{\textbf{Ablation of the reward weights $\lambda_{\text{ans}}$
and $\lambda_{\text{ent}}$} in \Cref{eq:skerl-reward} on
Qwen3-4B-Instruct, reported as overall \bench{} score. We constrain
$\lambda_{\text{ans}} + \lambda_{\text{ent}} = 1$ and sweep four splits.
The default setting used in \Cref{tab:skerl-main} is
$(\lambda_{\text{ans}}, \lambda_{\text{ent}}) = (0.7, 0.3)$.}
\label{tab:skerl-lambda-ablation}
\end{table}

The default $(0.7, 0.3)$ split is the peak across the four sweeps, and
performance degrades on both sides: shrinking the skill-entropy term to
$\lambda_{\text{ent}} = 0.1$ drops the \bench{} score by $7.6$ points,
while letting it dominate ($\lambda_{\text{ent}} \geq 0.5$) drops it by
$12.8$ to $19.8$ points. The asymmetric falloff is consistent with our
design intent: the answer signal must remain the dominant supervision,
and the skill-entropy reward is most useful as a structural shaping term
on top of it. Removing the entropy reward entirely is captured by the
GRPO baseline in \Cref{tab:skerl-main}, which sits below the default
split, so both endpoints of the simplex underperform $(0.7, 0.3)$.

\subsection{Comparison against the single-skill oracle}
\label{app:skerl_oracle}

\Cref{tab:skerl-main} reports cross-skill accuracy on \bench{}, where
the model receives the full task scenario together with all $L$
per-step questions. To check whether the gains of \skerl{} reflect a
genuine improvement in the model's underlying skill ability rather
than only better handling of the cross-skill format, we additionally
report the base model's accuracy under the Single-Skill setting of
\Cref{sec:evaluation}, which evaluates each step independently with
no surrounding scenario. \Cref{tab:skerl-oracle} reproduces
\Cref{tab:skerl-main} with this single-skill row appended for both
Qwen3-4B-Instruct and Qwen3-1.7B.

\begin{table}[h]
\centering
\small
\setlength{\tabcolsep}{2.2pt}
\renewcommand{\arraystretch}{0.95}
\resizebox{\textwidth}{!}{%
\begin{tabular}{l@{\hspace{5pt}} ccccc cccc c @{\hspace{1.1pt}}c}
\toprule
\multirow{2}{*}{\textbf{Method}}
& \multicolumn{6}{c}{\textbf{Verifiable Domains}} & \multicolumn{3}{c}{\textbf{Open-Ended Domains}}
&
& \multirow{2.1}{*}{\textbf{\bench{}}} \\
\cmidrule(lr){2-7} \cmidrule(lr){8-10}
& {\footnotesize\textbf{Math}} & {\footnotesize\textbf{Coding}} & {\footnotesize\textbf{Science}} & {\footnotesize\textbf{Planning}} & {\footnotesize\textbf{Logic}}
& {\footnotesize\makecell{\textbf{Info.}\\\textbf{Extraction}}}
& {\footnotesize\makecell{\textbf{Instruction}\\\textbf{Following}}}
& {\footnotesize\makecell{\textbf{Ctx.}\\\textbf{Retrieval}}}
& {\footnotesize\makecell{\textbf{Creative}\\\textbf{Writing}}}
& &  \multirow{0.1}{*}{\textbf{Performance}}\\
\midrule
\rowcolor{green!30}\multicolumn{12}{l}{\textit{\text{\# Qwen3-4B-Instruct}}} \vspace{3pt}\\
Base model
 & 14.0 & 26.3 & 34.5 & \phantom{0}5.9 & 12.9 & 47.8 & 64.2 & 59.2 & 68.8 & & 34.4 \\
\addlinespace[2pt]
SFT
 & 37.5 & 33.3 & 61.6 & 55.6 & 38.6 & 69.4 & 69.4 & 55.4 & 60.2 & & 55.8 \\
\addlinespace[2pt]
GRPO
 & 44.7 & 42.8 & 54.6 & \textbf{67.2} & 41.4 & 73.7 & 75.0 & 63.9 & 68.8 & & 58.8 \\
\addlinespace[2pt]
\textbf{Skill-Entropy RL}
 & \textbf{49.3} & \textbf{47.8} & \textbf{71.1} & 55.8 & \textbf{47.1} & \textbf{76.8} & \textbf{75.2} & \textbf{64.7} & \textbf{85.6} & & \textbf{68.4} \\
\addlinespace[2pt]
\textit{Single-skill oracle}
 & \textit{52.4} & \textit{49.5} & \textit{62.7} & \textit{15.2} & \textit{58.6} & \textit{70.5} & \textit{77.2} & \textit{71.3} & \textit{96.0} & & \textit{62.2} \\
\midrule
\rowcolor{yellow!25}\multicolumn{12}{l}{\textit{\text{\# Qwen3-1.7B}}} \vspace{3pt}\\
Base model
 & 11.7 & 13.0 & 13.3 & \phantom{0}9.4 & \phantom{0}9.4 & 17.7 & 16.5 & 14.1 & 17.8 & & 14.6 \\
\addlinespace[2pt]
SFT
 & 12.3 & 27.5 & 43.7 & 34.8 & 17.1 & 37.8 & \textbf{32.8} & 40.6 & 42.9 & & 30.6 \\
\addlinespace[2pt]
GRPO
 & 14.3 & 30.3 & 44.4 & 36.1 & 15.7 & 37.2 & 26.9 & 44.3 & 48.9 & & 32.2 \\
\addlinespace[2pt]
\textbf{Skill-Entropy RL}
 & \textbf{27.4} & \textbf{36.8} & \textbf{52.6} & \textbf{37.8} & \textbf{23.1} & \textbf{51.5} & 27.2 & \textbf{45.1} & \textbf{59.7} & & \textbf{40.1} \\
\addlinespace[2pt]
\textit{Single-skill oracle}
 & \textit{30.4} & \textit{41.5} & \textit{43.0} & \phantom{0}\textit{9.7} & \phantom{0}\textit{7.1} & \textit{52.5} & \textit{35.3} & \textit{28.2} & \textit{46.4} & & \textit{37.1} \\
\bottomrule
\end{tabular}}
\vspace{4pt}
\caption{\textbf{\skerl{} versus the single-skill oracle.} We extend
\Cref{tab:skerl-main} with a \textit{single-skill oracle} row
reporting the base model's accuracy when each step is evaluated
independently with no surrounding scenario (Single-Skill setting of
\Cref{sec:evaluation}). On both Qwen3-4B-Instruct and Qwen3-1.7B,
\skerl{} attains a higher overall \bench{} score in the cross-skill
setting than the base model achieves under the Single-Skill setting.}
\label{tab:skerl-oracle}
\end{table}

\subsection{Additional base models}
\label{app:skerl_additional_models}

\Cref{tab:skerl-additional-models} complements \Cref{tab:skerl-main}
with results on two additional base models, Llama-3.2-3B-Instruct and
Olmo3-7B-Instruct, using the same training data and configuration as
\Cref{sec:skerl_setup}. On both models, \skerl{} attains the highest
overall \bench{} score, matching the pattern observed on
Qwen3-4B-Instruct and Qwen3-1.7B in \Cref{tab:skerl-main}.

\begin{table}[h]
\centering
\small
\setlength{\tabcolsep}{2.2pt}
\renewcommand{\arraystretch}{0.95}
\resizebox{\textwidth}{!}{%
\begin{tabular}{l@{\hspace{5pt}} ccccc cccc c @{\hspace{1.1pt}}c}
\toprule
\multirow{2}{*}{\textbf{Method}}
& \multicolumn{6}{c}{\textbf{Verifiable Domains}} & \multicolumn{3}{c}{\textbf{Open-Ended Domains}}
&
& \multirow{2.1}{*}{\textbf{\bench{}}} \\
\cmidrule(lr){2-7} \cmidrule(lr){8-10}
& {\footnotesize\textbf{Math}} & {\footnotesize\textbf{Coding}} & {\footnotesize\textbf{Science}} & {\footnotesize\textbf{Planning}} & {\footnotesize\textbf{Logic}}
& {\footnotesize\makecell{\textbf{Info.}\\\textbf{Extraction}}}
& {\footnotesize\makecell{\textbf{Instruction}\\\textbf{Following}}}
& {\footnotesize\makecell{\textbf{Ctx.}\\\textbf{Retrieval}}}
& {\footnotesize\makecell{\textbf{Creative}\\\textbf{Writing}}}
& &  \multirow{0.1}{*}{\textbf{Performance}}\\
\midrule
\rowcolor{green!30}\multicolumn{12}{l}{\textit{\text{\# Llama-3.2-3B-Instruct}}} \vspace{3pt}\\
Base model
 & \phantom{0}2.0 & \phantom{0}4.3 & \phantom{0}4.9 & \phantom{0}3.1 & \phantom{0}0.0 & 10.8 & 17.1 & 10.5 & 14.6 & & \phantom{0}7.4 \\
\addlinespace[2pt]
SFT
 & 19.8 & 22.8 & 32.8 & 34.0 & \phantom{0}7.1 & 46.6 & 35.7 & 32.1 & 31.1 & & 33.2 \\
\addlinespace[2pt]
GRPO
 & 21.8 & 19.4 & 42.0 & 34.6 & \phantom{0}9.4 & 55.7 & 41.0 & 39.5 & 39.5 & & 37.8 \\
\addlinespace[2pt]
\textbf{Skill-Entropy RL}
 & \textbf{24.6} & \textbf{27.8} & \textbf{48.6} & \textbf{35.1} & \textbf{17.1} & \textbf{57.4} & \textbf{43.0} & \textbf{39.8} & \textbf{40.5} & & \textbf{40.7} \\
\midrule
\rowcolor{yellow!25}\multicolumn{12}{l}{\textit{\text{\# Olmo3-7B-Instruct}}} \vspace{3pt}\\
Base model
 & \phantom{0}3.4 & \phantom{0}6.5 & \phantom{0}8.8 & \phantom{0}0.8 & \phantom{0}0.0 & 10.4 & 24.5 & 16.1 & 38.3 & & 13.6 \\
\addlinespace[2pt]
SFT
 & 36.4 & 30.8 & 47.5 & 37.6 & \phantom{0}8.6 & 56.1 & 47.1 & 47.3 & \textbf{55.7} & & 46.0 \\
\addlinespace[2pt]
GRPO
 & 39.8 & \textbf{38.7} & 57.8 & 33.7 & 14.3 & 65.1 & 47.2 & 46.3 & 44.4 & & 49.1 \\
\addlinespace[2pt]
\textbf{Skill-Entropy RL}
 & \textbf{41.3} & 35.8 & \textbf{59.8} & \textbf{48.3} & \textbf{14.9} & \textbf{68.3} & \textbf{51.9} & \textbf{47.6} & 51.2 & & \textbf{52.2} \\
\bottomrule
\end{tabular}}
\vspace{4pt}
\caption{\textbf{Additional base models.} Per-domain accuracy and
overall \bench{} score on Llama-3.2-3B-Instruct and Olmo3-7B-Instruct,
under the same training data and configuration as \Cref{tab:skerl-main}.
\skerl{} attains the highest overall \bench{} score on both models.}
\label{tab:skerl-additional-models}
\end{table}

\subsection{Generalization to external benchmarks}
\label{app:skerl_external_benchmarks}

We evaluate the Base, GRPO, and \skerl{} checkpoints from
\Cref{tab:skerl-main} on five external benchmarks that are neither
generated by nor used in our pipeline: two long-horizon reasoning
benchmarks (MuSR and LongBench-MuSiQue) and three general reasoning
benchmarks (GPQA-Diamond, MMLU, and IFEval). All inference-time
settings follow \Cref{app:eval_settings}. \Cref{tab:skerl-external}
reports per-benchmark accuracy together with the overall average
across the five benchmarks.

\begin{table}[h]
\centering
\small
\setlength{\tabcolsep}{6pt}
\renewcommand{\arraystretch}{1}
\begin{tabular}{l@{\hspace{5pt}} ccccc c @{\hspace{1.1pt}}c}
\toprule
\textbf{Method}
 & {\footnotesize\textbf{MuSR}}
 & {\footnotesize\makecell{\textbf{LongBench-}\\\textbf{MuSiQue}}}
 & {\footnotesize\makecell{\textbf{GPQA-}\\\textbf{Diamond}}}
 & {\footnotesize\textbf{MMLU}}
 & {\footnotesize\textbf{IFEval}}
 & & \textbf{Avg.}\\
\midrule
\rowcolor{yellow!25}\multicolumn{8}{l}{\textit{\text{\# Qwen3-1.7B}}} \vspace{3pt}\\
Base model
 & 37.4 & 12.3 & \phantom{0}5.6 & 23.1 & \textbf{33.8} & & 22.4 \\
\addlinespace[2pt]
GRPO
 & 39.8 & \textbf{21.9} & \phantom{0}4.0 & 23.1 & 32.3 & & 24.2 \\
\addlinespace[2pt]
\textbf{Skill-Entropy RL}
 & \textbf{40.7} & 21.8 & \phantom{0}\textbf{6.1} & 23.1 & 33.1 & & \textbf{25.0} \\
\midrule
\rowcolor{green!30}\multicolumn{8}{l}{\textit{\text{\# Qwen3-4B-Instruct}}} \vspace{3pt}\\
Base model
 & 40.7 & 23.8 & \phantom{0}7.6 & 63.1 & 82.8 & & 43.6 \\
\addlinespace[2pt]
GRPO
 & 40.8 & 28.8 & \textbf{10.1} & 68.1 & 82.2 & & 46.0 \\
\addlinespace[2pt]
\textbf{Skill-Entropy RL}
 & \textbf{41.8} & \textbf{29.1} & \textbf{10.1} & \textbf{69.3} & \textbf{83.0} & & \textbf{46.7} \\
\bottomrule
\end{tabular}
\vspace{4pt}
\caption{\textbf{Generalization to external benchmarks.} Per-benchmark
accuracy and overall average on five external benchmarks not used in
our pipeline. \skerl{} attains the highest average at both model
sizes.}
\label{tab:skerl-external}
\end{table}

We make three observations. (1) \emph{Long-horizon transfer.}
\skerl{} attains the best MuSR score at both scales and the best
LongBench-MuSiQue score on Qwen3-4B-Instruct, indicating that the
skill-entropy reward extends to multi-step reasoning benchmarks
outside our pipeline. (2) \emph{General reasoning.} \skerl{} attains
the highest average score across the five benchmarks at both scales,
matching or outperforming GRPO on all but one entry, so the
skill-entropy reward does not compromise general reasoning
performance. (3) \emph{No forgetting.} \skerl{} matches or exceeds
the base model on almost every benchmark for both models. These
observations complement \Cref{tab:skerl-openr1}, where the same
reward applied to off-the-shelf OpenR1-Math data outperforms GRPO on
all six external math benchmarks by $+1.9$ on average, providing
additional evidence that the skill-entropy reward generalizes to
downstream tasks.

\subsection{Worked example: labeling an off-the-shelf problem}
\label{app:skerl_offshelf_example}

We illustrate the labeling pipeline of \Cref{sec:skerl_plugin} on a
single OpenR1-Math problem. Given the (problem, gold trace) pair, the
annotator (Qwen3-8B) segments the trace into reasoning steps, assigns
each step a (domain, skill) label drawn from the \bench{} skill bank
(\Cref{app:skill_bank}), and emits the per-step intermediate conclusion
as the step's gold answer. The resulting skill sequence is fed into
\Cref{eq:task_entropy} to give the task-level skill entropy.

\begin{examplelisting}{Worked example: labeling an OpenR1-Math problem}
[Problem]
Let f(x) = x^3 - 6x^2 + 11x - 6. Find the sum of all real roots of
f(f(x)) = 0.

[Gold trace, segmented and labeled]
Step 1 -- skill = (math, factoring)
  Factor f(x) = (x-1)(x-2)(x-3), so f(x) = 0 has roots {1, 2, 3}.
  intermediate answer: roots of f are 1, 2, 3.

Step 2 -- skill = (math, polynomial_equations)
  f(f(x)) = 0 iff f(x) in {1, 2, 3}; solve f(x) = k for each k by
  shifting the constant term.
  intermediate answer: nine real roots, three from each cubic
  f(x) - k = 0.

Step 3 -- skill = (math, vieta_formulas)
  Each cubic f(x) - k = x^3 - 6x^2 + 11x - (6+k) has root sum 6 by
  Vieta. Total root sum = 3 * 6 = 18.
  final answer: 18.

[Resulting skill sequence and entropy]
skill sequence: (math, factoring) -> (math, polynomial_equations)
                -> (math, vieta_formulas)
task-level skill entropy: computed from the pairwise table of Sec. 3.1.
\end{examplelisting}

The same procedure runs over the full 6K subset of OpenR1-Math used
in \Cref{sec:skerl_plugin}, balanced across the three skill-entropy
levels and matched in size to the \bench{} RL split. Labels that fall
outside the skill bank are routed to the nearest canonical key by the
embedding lookup of \Cref{app:skerl_skill_match}, so the labeling
vocabulary stays identical to the one used to compute the pairwise
skill-entropy table. Baselines for \Cref{tab:skerl-openr1} are the
base model, SFT on the same labeled traces, and GRPO trained with
the answer reward alone (the ablation of our method without the
skill-entropy reward).


\end{document}